\documentclass[journal]{IEEEtran}
\usepackage[bookmarks=true,bookmarksnumbered=true,bookmarkstype=toc, %
pdftitle={Stiffness Optimization for Concentrated Bending in Magnetically Actuated Catheters},
pagebackref=false,
colorlinks,
urlcolor=black,
citecolor=magenta,
linkcolor=blue,
anchorcolor=red
]{hyperref}
\usepackage[linesnumbered, ruled]{algorithm2e}
\SetKwRepeat{Do}{do}{while}%
\usepackage{cite}
\usepackage{comment}
\usepackage{amsmath,amssymb,amsfonts}
\usepackage{graphicx}
\usepackage{textcomp}
\usepackage{xcolor}
\usepackage{color}
\usepackage{float}
\usepackage{amsmath,bm}
\usepackage{gensymb}
\usepackage{stfloats}
\usepackage{subfigure}
\usepackage{booktabs}
\usepackage{threeparttable}
\usepackage{pifont}
\usepackage{siunitx}
\usepackage{array}
\usepackage{amsmath}
\usepackage{subfigure}
\usepackage{makecell}
\usepackage[ruled,linesnumbered]{algorithm2e}
\usepackage{bm}
\usepackage{textcomp,gensymb}
\usepackage{tikz,xcolor,hyperref}

\definecolor{lime}{HTML}{A6CE39}
\DeclareRobustCommand{\orcidicon}{
\begin{tikzpicture}
\draw[lime, fill=lime] (0,0)
circle[radius=0.16]
node[white]{{\fontfamily{qag}\selectfont \tiny \.{I}D}}; 
\end{tikzpicture}
\hspace{-2mm}
}
\foreach \x in {A, ..., Z}{%
\expandafter\xdef\csname orcid\x\endcsname{\noexpand\href{https://orcid.org/\csname orcidauthor\x\endcsname}{\noexpand\orcidicon}}
}

\begin{document}
\title{Stiffness Optimization for Concentrated Bending in Magnetically Actuated Catheters: Maintaining Steerability under Gradient Stiffness}

\author{Jiewen Tan\orcidA{}, Junnan Xue\orcidB{}, Shing Shin Cheng\orcidD{}, Shuang Song\orcidE{}, Erli Lyu\orcidF{} and Jiaole Wang\orcidG{} \\%
		\thanks{This work was supported partly by National Key R\&D Program of China under Grant 2022YFB4703200, and partly by Talent Recruitment Project of Guangdong under Grant 2021QN02Y839, and in part by the Science Technology Innovation Committee of Shenzhen under Grant JCYJ20220818102408018 and GXWD20231129103418001. Corresponding authors: Jiaole Wang (wangjiaole@hit.edu.cn), Shuang Song (songshuang@hit.edu.cn).}
		\thanks{Jiewen Tan, Junnan Xue, Shuang Song and Jiaole Wang are with Harbin Institute of Technology (Shenzhen), Shenzhen, China, 518055. %
		Shing Shin Cheng is with the Department of Mechanical and Automation Engineering and T Stone Robotics Institute, The Chinese University of Hong Kong, Hong Kong.
        Erli Lyu is with Faculty of Applied Science, Macao Polytechnic University, Macao.}
        \thanks{This work has been submitted to the IEEE for possible publication. Copyright may be transferred without notice, after which this version may no longer be accessible.}}%
\maketitle
	
\begin{abstract}
Achieving both efficient pushability (propulsion transmission) and proximally concentrated bending for steerability is challenging for magnetically actuated soft catheters: higher axial/bending stiffness improves force transmission but reduces steerability, whereas lower stiffness enables large, proximally concentrated bending yet increases kinking/buckling risk under compressive push loads.
To address this trade-off, we propose a stiffness-optimized multi-segment magnetically actuated catheter (SO-MAC) that integrates a decoupled steering-advancement mechanism with a gradient-stiffness architecture.
The SO-MAC concentrates bending about a stable proximal pivot during advancement while the distal section passively self-straightens to transmit propulsion, aided by the optimized stiffness distribution and elastic recovery of the spring backbone against friction-induced kinking/buckling.
Over $0{-}180^{\circ}$ combined steering and advancement, the pivot remained stable and the distal tip advanced near-straight toward the target direction.
A 1.5 mm-diameter SO-MAC achieved up to $180^{\circ}$ steering with a 3 mm bending radius at its 10~mm tip, with an average shape error of $1.39 \pm 0.56$~mm and a steering-pivot error of $0.35 \pm 0.10$~mm.
Visual feedback control in a bronchial phantom further confirmed robust navigation through highly curved, bifurcating paths.
\end{abstract}

\section{Introduction}
\label{sec:intro}
\IEEEPARstart{S}{urgical} catheter robots have become essential tools for minimally invasive surgery, including cable‑driven robots \cite{wang2023novel,zhang2024AI}, eccentric‑tube robots \cite{wang2021eccentric}, and fluid‑driven robots \cite{ZHANG2024oct,zhang2025kinematic}. 
These approaches have contributed to significant advances in miniaturization and dexterity \cite{Ultrasound2025XUtiantian}.
Among these options, magnetically actuated catheters (MACs) offer unique advantages: they eliminate pre‑tension forces and complex transmission structures, while reducing the risks of air or fluid leakage. 
Such inherent features of MACs facilitate further miniaturization and enable safe navigation through intricate anatomical pathways \cite{yuan2024motor}.
MACs have been applied in various minimally invasive procedures \cite{Magnetic2025Cao, wang2021endoscopy, wang2025magnetically,fu2025multifunctional,tan2023model}, including vascular interventions \cite{fu2023magnetically,yang2021magnetic,Automatic2024Xutiantian}, transesophageal echocardiography \cite{Closed2023Keyu}, gastrointestinal diagnostics \cite{xu2022adaptive}, and bronchoscopy \cite{yuan2024motor,yuan2025endotracheal}. 

MACs are steered by applying a torque to their magnetic section through controlled orientation changes of the external magnetic field.
Early designs achieved this function by attaching a small permanent magnet at the distal end of a compliant polymer or composite shaft \cite{chun2007remote,wang2021Kinematic}. 
Subsequent generations distributed multiple magnets along the catheter body to enhance the overall magnetic response \cite{edelmann2017magnetic}, while later developments embedded magnetic microparticles \cite{kim2019ferromagnetic} within soft elastomers to create magnetized flexible segments \cite{Magnetic2025Cao}. 
State-of-the-art microparticle-based MACs reach only about $80^{\circ}$ at $18\,\mathrm{mT}$~\cite{chen2025rapid}.
More recently, hybrid actuation strategies combining magnetic steering with shape‑memory alloys (SMAs) have been introduced to enable in‑situ programmable magnetization for shape reconfiguration \cite{Xue2025Reprogramming}, and with low‑melting‑point alloys (LMPAs) to achieve variable stiffness for compliance control \cite{lussi2021submillimeter}.

Among existing MAC designs, magnetic ball-chain structures exhibit remarkable steerability owing to their extremely low effective bending stiffness while retaining high magnetic content compared with particle-doped designs \cite{pittiglio2023magnetic}. 
In bifurcating channels, most of the bending can be concentrated to only a few balls negotiating the corner, with the remaining section staying relatively straight \cite{pittiglio2023magnetic}. We refer to this corner-concentrated deformation pattern as \emph{concentrated bending}. 
Such a concentrated-bending effect can facilitate bifurcation entry in branched luminal networks (e.g., airways) by enabling a near-straight distal approach into sharply angled child branches despite strong proximal confinement in a narrow parent lumen.
However, interventional navigation also demands axial force transmission (pushability) and resistance to kinking \cite{lawrence2005materials}, which are commonly addressed using distal-to-proximal stiffness gradients \cite{yang2025degradable,dreyfus2024dexterous}. 
Existing stiffness gradients are largely empirical and not optimized for magnetic actuation, leaving the stiffness-concentrated-bending relation and feasible stiffness limits unclear. 
Moreover, while increased stiffness can suppress friction-induced buckling under axial compressive push loads, excessive stiffness can impede pushability at large steering angles even without friction.

\begin{figure*}[t]
  \centering
  \includegraphics[width=0.95\textwidth]{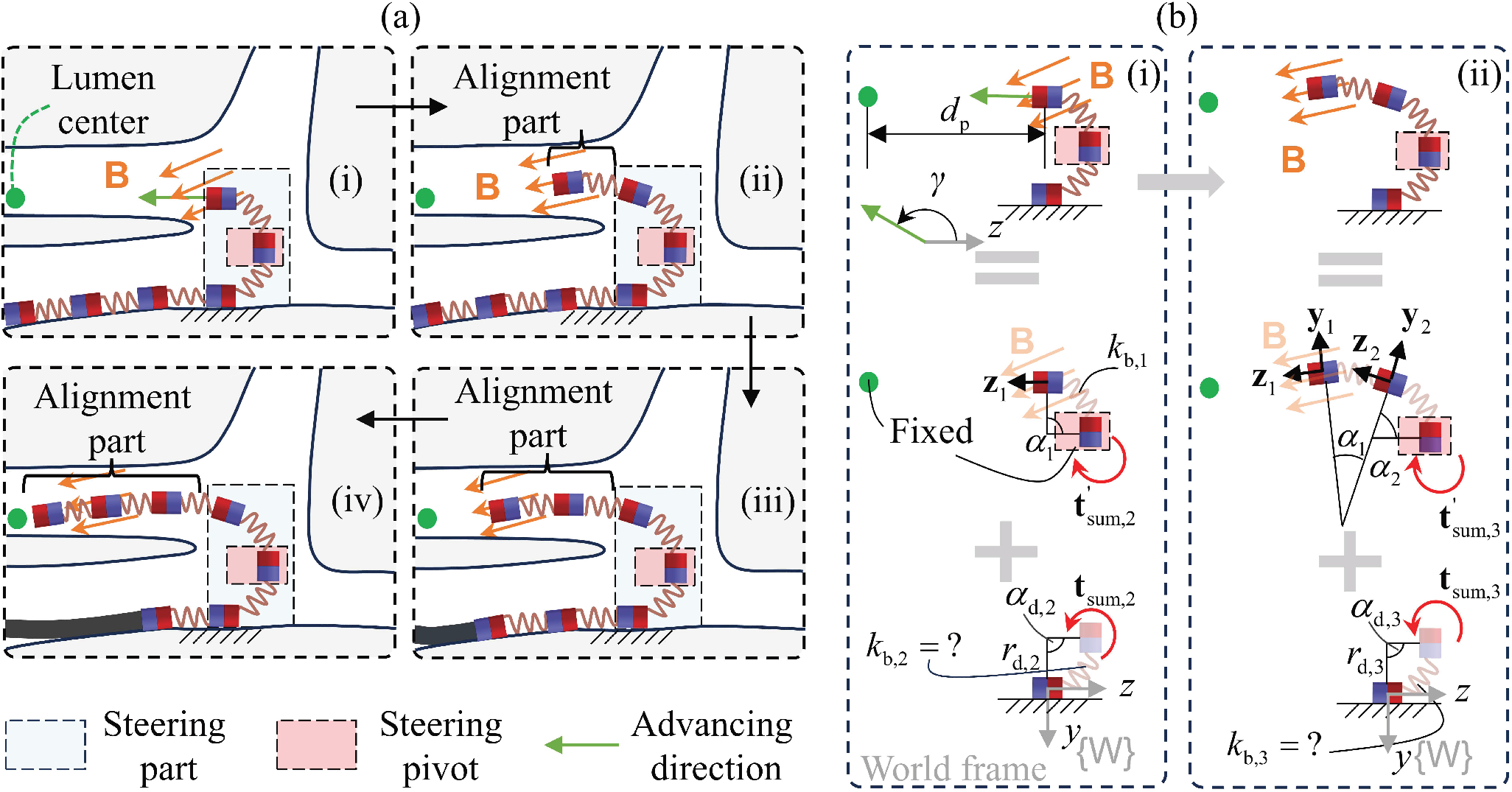}
  \caption{SO-MAC motion pattern towards a 180$^\circ$ target lumen.
  (a) (i)–(iv) show the SO-MAC navigating a 180$^\circ$ lumen, using the same steering pivot as a stable support during the steering phase.
  (b) The second and third bending stiffnesses, ${k_{{\rm{b,2}}}}$ and ${k_{{\rm{b,3}}}}$, are calculated using Hooke's law.
  ${\alpha _n}$ is the central angle of the $n$-th segment.
  The SO-MAC is divided into two segments from the steering pivot via equivalent transformation, with corresponding force analysis presented.
${\bf{t}}_{{\rm{sum,}}n}^\prime$ represents the net torque required to maintain the steering pivot in equilibrium and its position unchanged when ${{\bf{z}}_1}$ is aligned with the distant lumen center, while ${{\bf{t}}_{{\rm{sum,}}n}}$ denotes the reaction torque opposing ${\bf{t}}_{{\rm{sum,}}n}^\prime$.
  }
  \label{fig.motion_pattern}
 \end{figure*}

To address this gap, we propose a stiffness-optimized MAC (SO-MAC) that realizes decoupled steering and advancement via concentrated bending while maintaining a gradual stiffness increase. 
The SO-MAC comprises alternating magnetic and spring segments: the springs define an optimized local stiffness distribution that concentrates bending in the steering part and, via intrinsic elastic recoil, bias the sections outside the steering part (alignment part) to remain nearly straight even under axial compressive loads and frictional contact, improving pushability. 
Under the worst-case narrow-parent-lumen constraint (base clamped), the free section splits into a steering part that forms a fixed pivot (Fig.\ref{fig.motion_pattern}(a)) and an alignment part that aligns to the target-lumen center (green dot) and transmits propulsion over a 0$^\circ$--180$^\circ$ steering range.
The recursive stiffness design in Fig. \ref{fig.motion_pattern}(b) maintains the pivot location, producing an increasing stiffness distribution and stable steering.
This decoupled behavior is consistent with the boundary conditions assumed in the steering‑pivot model under worst‑case constraint.
The proposed kinetostatic model and fabrication method were experimentally validated, demonstrating accurate steering‑pivot control, preserved steerability and pushability across 0$^\circ$–180$^\circ$.

\section{Steering Pivot and Problem Formulation}
\label{sec:design}
The SO‑MAC tip, consisting of the first two segments, bends within 0°–180° to align with the lumen center, thereby defining the advancing direction (Fig.~\ref{fig.motion_pattern}(a) (i)). 
To ensure pushability and steerability, the steering pivot is assumed fixed while the alignment part straightens along the lumen centerline.
Simulations and experiments later confirmed that this design maintained stable steering‑pivot behavior for $\gamma$ ranging from 0° to 180°. 
In this configuration, the first two segments bend 180° with desired angles ${\alpha_1} = {\alpha_2} = \gamma /2=\pi/2$, and each spring—indexed by $n$—is assigned a designed curvature radius ${r_{{\rm d},n}}$ at the designed central angle ${\alpha_{{\rm d},n}} = \pi/2$ to establish the pivot (Fig.~\ref{fig.motion_pattern}(b)). The distance from the first magnet to the lumen center, $d_{\rm p}$, is fixed at 80 mm—the longest straight branch in the bronchial phantom (Section \ref{Navigation})—to keep the alignment part nearly straight and ensure effective advancement, where the external magnetic field (EMF) for steering is denoted as $\mathbf{B}$.

Given the first‑spring stiffness ${k_{\mathrm{b,1}}}$, the subsequent stiffnesses ${k_{\mathrm{b,2}}}$ and ${k_{\mathrm{b,3}}}$ are determined as shown in Fig.~\ref{fig.motion_pattern}(b). A kinetostatic model defines the equilibrium among the EMF, embedded magnets, and springs. 
For $t$ unsupported segments, the model yields a configuration vector $\bm{\alpha }\vert_{(t-1) \times 1} = [\alpha_1 \ \alpha_2 \ \cdots \ \alpha_{t-1}]^{\rm{T}}$, excluding the steering‑pivot segment (i.e., the $t$-th segment). Solving the equilibrium equations provides the resultant torque ${\mathbf{t}}_{{\rm{sum,}}n}^{\prime}$ required to maintain the fixed pivot when the magnetic field aligns the alignment part with the lumen center. The stiffness of the next steering‑pivot spring, ${k_{{\rm{b,}}n}}$ can be obtained by ${k_{{\rm{b,}}n}} = \left\| {{{\bf{t}}_{{\rm{sum,}}n}}} \right\|/{\alpha _{{\rm{d},}n}}$.

\section{Methods}%
\label{sec:setup}
Because axial compression from adjacent-magnet attraction is negligible compared with the spring compression stiffness, the spring length ${l_{\mathrm{s}}}$ is assumed constant. 
Magnets are modeled as dipoles obeying the inverse-cube law \cite{petruska2012optimal,wang2021Kinematic} and are assumed to experience primarily magnetic torques; adjacent-magnet forces are included only via equivalent torques. 
Each spring is approximated as an arc under transverse bending moments \cite{wahl1944mechanical}, and only nearest-neighbor and external-field interactions are considered.

\subsection{Kinetostatic Modeling}
\label{Kinematic Modeling}
Due to geometric symmetry, the kinetostatic model is restricted to the left y–z half‑plane of the world frame (Fig.~\ref{fig.motion_pattern}(b)), as the transition from a parent lumen to a child lumen can be approximated as an in‑plane deflection. 

\begin{figure}[t]
  \centering
  \includegraphics[ width=0.45\textwidth]{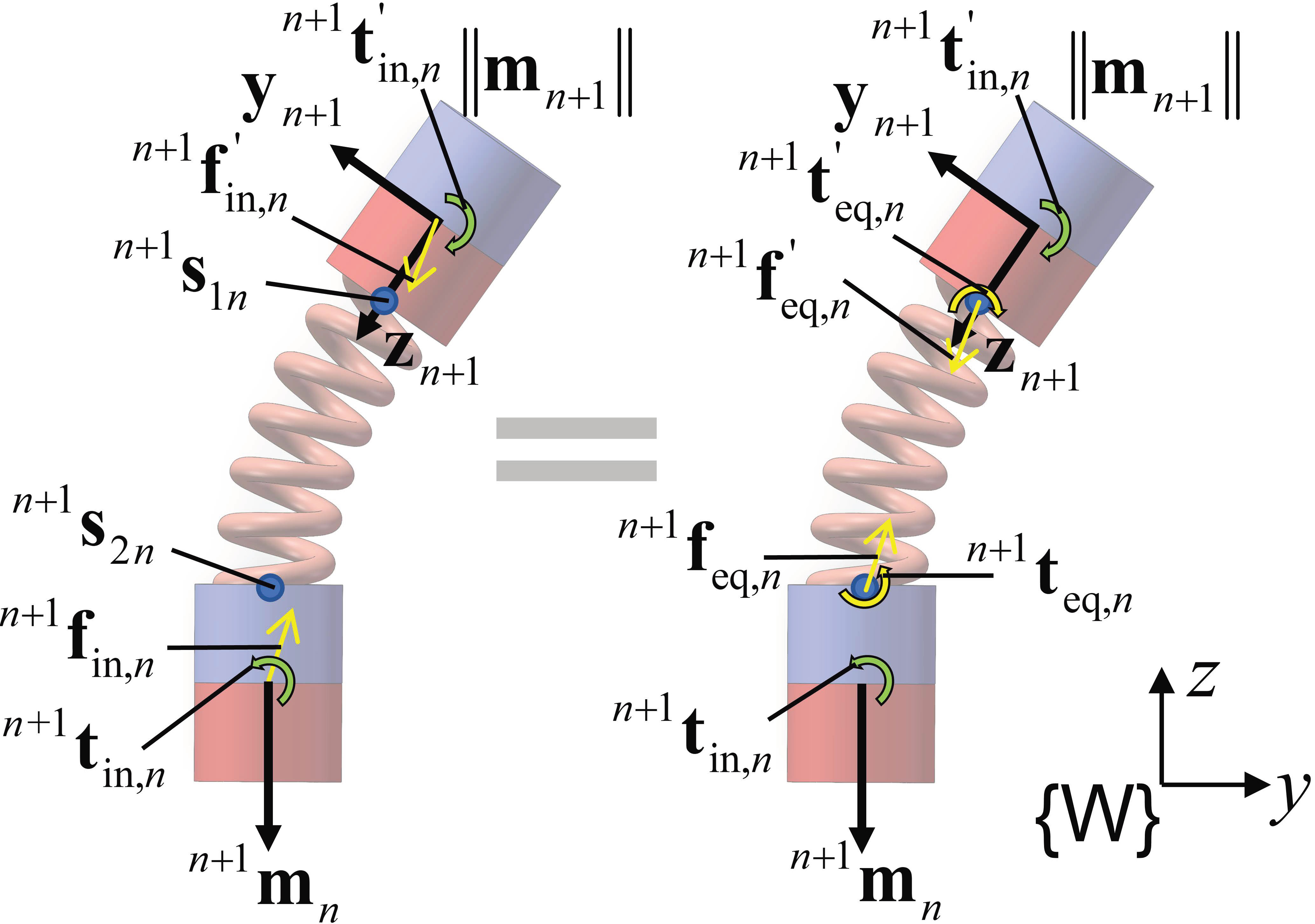}
  \caption{Equivalent static analysis.
  The attractive force between adjacent magnets induces a static equilibrium in the spring, equivalent to a pair of forces and moments applied at its two ends. 
  This is also a similar method for analyzing the forces on springs\cite{wang2023novel}.}
  \label{fig.euqi}
 \end{figure}

The central angle of the $n$-th spring is ${\alpha _{n}}$, where $n = 1, 2, \cdots, t$, and $t$ is the number of unsupported segments in SO-MAC.
${l_\mathrm{m}}$ is the length of the magnets of SO-MAC.
As shown in Fig. \ref{fig.motion_pattern}(b), we first discuss the transformation matrix of a single spring joint.
${r_{{\rm{d,}}n}}$ is the desired radius of curvature for the $n$-th spring at ${\alpha _{{\rm{d}},n}} = {\pi/2}$.
The constant spring length $l_{\rm{s}} = r_{{\rm{d,}}n}\alpha_{{\rm{d},}n}$ is the same for each joint.
Each spring $n$ corresponds to a radius of curvature $r_{n}=\frac{l_{\rm{s}}}{\alpha_n}$, a bending stiffness $k_{b,n}$ and a compression stiffness $k_{c,n}$.
In the $n$-th magnet-spring segment under the current frame of the $\left( {n + 1} \right)$th magnet, as shown in Fig. \ref{fig.euqi}, the positions of the two ends of spring $n$ are defined as ${}^{n + 1}{{\bf{s}}_{1n}} = {[0,\;0,\;\frac{{{l_{\rm{m}}}}}{2}]^{\rm{T}}}$ and ${}^{n + 1}{{\bf{s}}_{2n}} = {[0,\; - {r_n}(1 - \cos {\alpha _n}),\;\frac{{{l_{\rm{m}}}}}{2} + {r_n}\sin {\alpha _n}]^{\rm{T}}}$.
The position of the $n$-th magnet is also defined in this frame as
\begin{equation}
    {}^{n + 1}{\mathbf{r}_{{\rm{d}},n}} = \left[ {\begin{array}{*{20}{c}}
    0\\
    { - {r_n}\left( {1 - \cos {\alpha _n}} \right) - \frac{{{l_{\rm{m}}}}}{2}\sin {\alpha _n}}\\
    {\frac{{{l_{\rm{m}}}}}{2}\left( {1 + \cos {\alpha _n}} \right) + {r_n}\sin {\alpha _n}}
    \end{array}} \right].
    \label{p1}
\end{equation}
${\mathbf{s}_{1n}}$ represents the central point of one end of the spring, positioned near the $(n+1)$th magnet. ${\mathbf{s}_{2n}}$ denotes the central point of the other end of the spring, situated near the $n$th magnet.
The rotation matrix of the $n$th magnet under the $(n+1)$th magnet is 
\begin{equation}
\begin{split}
    ^{n + 1}{{\mathbf{R}}_n} &= \left[ {\begin{array}{*{20}{c}}
1&0&0\\
0&{\cos {\alpha _n}}&{ - \sin {\alpha _n}}\\
0&{\sin {\alpha _n}}&{\cos {\alpha _n}}
\end{array}} \right] \\
\end{split}
    \label{rotation}
\end{equation}
Hence, we have the transformation matrix from the $n$-th magnet to the $(n+1)$-th magnet is
\begin{equation}
    {}^{n + 1}{{\mathbf{T}}_n} = \left[ {\begin{array}{*{20}{c}}
    {{{}^{n + 1}{\mathbf{R}}_n}}&{{{}^{n + 1}{\mathbf{r}}_{{\rm{d,}}n}}}\\
    {{{\mathbf{o}}_{1 \times 3}}}&1
    \end{array}} \right].
\end{equation}

Then, if the SO-MAC has $t$ unsupported segments, the pose matrix of each magnet $n$ relative to the fixed world frame $\{{\rm{W}}\}$ (as defined in Fig.\ref{fig.motion_pattern}(b) (i)) can be expressed as 
\begin{equation}
    {}^{\rm{w}}{{\mathbf{T}}_n} = \prod\limits_{k = n}^t {{{}^{(t+n-k)+1}{\mathbf{T}}_{t + n - k}}}  = 
    \begin{bmatrix}
    {{{\mathbf{x}}_n}}&{{{\mathbf{y}}_n}}&{{{\mathbf{z}}_n}}&{{{\mathbf{p}}_n}}\\
    0&0&0&1
    \end{bmatrix},
    \label{position_allmag}
\end{equation}
where ${{{\mathbf{z}}_n}}$ and ${{{\mathbf{p}}_n}}$ are the magnetic moment direction and position of the $n$-th magnet.

According to \cite{petruska2012optimal}, the magnetic field ${{\mathbf{B}}_{n + 1}}$ generated by the $\left( {n + 1} \right)-$th magnetic dipole moment ${{\mathbf{m}}_{n+1}}$ at ${{\mathbf{m}}_n}\in \mathbb{R}^{3}$ can be expressed as

\begin{equation}
\begin{split}
    {{\mathbf{B}}_{n + 1}}\left( {{{}^{n+1}{\mathbf{r}}_{{\rm{d,}}n}},{{\mathbf{m}}_{n + 1}}} \right)= ~~~~~~~~~~~~~~~~~~~~~~~~~~~~~~~~~~~~~\\ \frac{{{\mu _0}\left\| {{{\mathbf{m}}_{n + 1}}} \right\|}}{{4\pi {{\left\| {{{{}^{n + 1}\mathbf{r}}_{{\rm{d,}}n}}} \right\|}^3}}} \left( {3({{{}^{n + 1}{\mathbf{\hat r}}}_{{\rm{d,}}n}}{{}^{n + 1}{{\mathbf{\hat r}}}^{\rm{T}}}_{{\rm{d,}}n}) - {\mathbf{I}}} \right){{\mathbf{\hat m}}_{n + 1}},
\end{split}
\end{equation}
where $\hat{\cdot}$ denotes unit vector, and $\left\|  \cdot  \right\|$ refers to the L2 norm of any vector.
We used identical permanent magnets which had the same magnitude of magnetic dipole intensity $\left\| {{{\mathbf{m}}_n}} \right\| = {B_r}{V_{\rm{m}}}/{\mu _0}$, where ${V_{\rm{m}}}$ is the volume of a magnet, ${\mu _0} \approx 4\pi  \times {10^{ - 7}}{\rm{N/}}{{\rm{A}}^{\rm{2}}}$ is the vacuum permeability and ${{B_r}}$ is the remanence of the magnets.
Magnet $n$ is subjected to the magnetic force ${}^{n+1}{{\mathbf{f}}_{{\rm{in,}}n}}\in \mathbb{R}^{3}$ and magnetic moment ${}^{n+1}{{\mathbf{t}}_{{\rm{in,}}n}}\in \mathbb{R}^{3}$ exerted by magnet $(n+1)$ as 
\begin{equation}
    \begin{array}{l}
    {{}^{n+1}{\mathbf{f}}_{{\rm{in,}}n}} = ({}^{n+1}{{\mathbf{m}}_n}\cdot \nabla ){{\mathbf{B}}_{n + 1}}\\
    {{}^{n+1}{\mathbf{t}}_{{\rm{in,}}n}} = {{}^{n+1}{\mathbf{m}}_n} \times {{\mathbf{B}}_{n + 1}}
    \end{array},
    \label{inner}
\end{equation}
where ${}^{n + 1}{{\bf{m}}_n} = \left\| \bf{m}_n \right\| \, {}^{n + 1}{{\bf{z}}_n}$ is the magnetic moment of magnet $n$ under the $\{n+1\}$ frame.

We can apply an equivalent transformation to the inner magnetic forces, ${}^{n + 1}{{\mathbf{f}}_{{\rm{in,}}n}}$ and ${}^{n + 1}{\mathbf{f}}_{{\rm{in,}}n}^{'}$, to obtain the corresponding equivalent magnetic torques ${}^{n + 1}{\mathbf{t}}_{{\rm{eq,}}n}$ they contribute.
After the equivalent transformation in Fig. \ref{fig.euqi}, we have
\begin{equation}
    \begin{array}{l}
    {{}^{n + 1}{\mathbf{f}}_{{\rm{eq,}}n}} = {{}^{n + 1}{\mathbf{f}}_{{\rm{in,}}n}}\\
    {{}^{n + 1}{\mathbf{t}}_{{\rm{eq,}}n}} = {{}^{n + 1}{\mathbf{f}}_{{\rm{in,}}n}} \times ({{}^{n + 1}{\mathbf{s}}_{2n}} - {{}^{n + 1}{\mathbf{r}}_{{\rm{d,}}n}}).
    \end{array}
\end{equation}

We consider there are $t$ movable springs.
All the torques acting on the $n$-th spring consist of the magnetic torques ${{\mathbf{t}}_i} = \left\| {{{\mathbf{m}}_i}} \right\|{{\mathbf{z}}_i} \times {\mathbf{B}}~(i=1,2,\cdots,n)$ exerted by the EMF (${\mathbf{B}}\in \mathbb{R}^{3}$) on the $n$ magnets, as well as the torques ${\mathbf{t}}_{{\rm{eq,}}n}$ and ${\mathbf{t}}_{{\rm{in,}}n}$, both expressed in the world frame, exerted by the $(n+1)$-th magnet on the $n$-th magnet.
The internal torques ${\mathbf{t}}_{{\rm{eq}}}$ and ${\mathbf{t}}_{{\rm{in}}}$ from the first to the $\left( {n - 1} \right)$-th spring are action-reaction moment pairs, so they have no effect on the $n$-th spring.
According to \cite{wahl1944mechanical}, we have
\begin{equation}
 {{\mathbf{t}}_{{\rm{sum,}}n}} = \left\| {\sum\limits_{i = 1}^n {{{\mathbf{t}}_i}}  + {{\mathbf{t}}_{{\rm{eq}},n}} + {{\mathbf{t}}_{{\rm{in}},n}}} \right\| = {k_{{\rm{b}},n}}{\alpha _n} .
    \label{non_equa}
\end{equation}

\subsection{Solving Approach of Kinetostatic Model}
\label{Solving Approach of Forward Kinetostatic Model}

For given ${\mathbf{B}}$ and ${k_{{\rm{b,}}n}}$, we solve (\ref{non_equa}) for ${\bm{\alpha} \vert_{t \times 1}} \in \mathbb{R}{^{t}}$; for numerical stability (some ${\alpha_n}\to 0$ when $\theta>0$ and as the SO-MAC length increases), we scale (\ref{non_equa}) by a scale factor $s_n$ as
\begin{equation}
    s_{n} \left( {\left\| {\sum\limits_{i = 1}^n {{{\mathbf{t}}_i}}  + {{\mathbf{t}}_{{\rm{eq}},n}} + {{\mathbf{t}}_{{\rm{in}},n}}} \right\| - {k_{{\rm{b}},n}}{\alpha _n}} \right) = 0,
    \label{normEQ}  
\end{equation}
where 
\begin{equation}
    {s_n} = {10^{ - floor\left( {\lg \left( {\left\| {\sum\limits_{i = 1}^n {{{\mathbf{t}}_i} + {{\mathbf{t}}_{{\rm{eq,}}n}} + {{\mathbf{t}}_{{\rm{in,}}n}}} } \right\|} \right)} \right)}}.
\end{equation}

\subsection{Bending Stiffness Design}
\label{BDesign}
To enable the SO-MAC to advance toward $\gamma=180^\circ$ while maintaining steering pivot, the bending stiffness of each spring segment ${k_{{\rm{b}},n}}$ must be designed. 
The calculation of ${k_{{\rm{b}},n}}$ is divided into two cases: $n = 2$ and $n \ge 3$. In both cases, we set $\alpha_n=\alpha_{\rm{d},n} = \pi/2$ to maintain the steering pivot position and define the unsupported segment number as $t = n$.
For $n = 2$, given ${k_{{\rm{b}},1}}$ and the magnetic field direction $\theta$, the net torque acting on the second spring can be directly computed. 
For $n \ge 3$, the net torque on the $n$-th spring is obtained by solving for the SO-MAC configuration and $\theta$—subject to the constraint (\ref{non_equa})—when the alignment part aligns with the target lumen center. 
%
The magnetic field acting on the SO-MAC has a bounded strength.
${\mathbf{\left\| {B} \right\|}}=40~\rm{mT}$ is our design reference since it is within the range of the similar system \cite{dreyfus2024dexterous}. 

\subsubsection{$n=2$}
We design the bending stiffness based on the most challenging scenario, where the target lumen direction $\gamma$ is $180^\circ$.
By rotating the external magnetic field ${\mathbf{B}}$, we aim to transform the first two springs of the SO-MAC from a straight configuration to the desired state, achieving a tip bending angle of ${\gamma_{\mathrm{max}}} = \alpha_1^* + \alpha_2^* = \pi$, with a designed radius of curvature ${r_{\mathrm{d},n}}$ for the two springs.
Since $\alpha_2^*=\alpha_{{\rm{d}},2}=\pi/2$, $\alpha_1^*=\gamma_{\rm{max}}-\alpha_2^*=\pi/2$.
$\alpha_1^*$ and $\alpha_2^*$ denote the numerical solutions for the first two springs.
To achieve $\gamma_{\rm{max}}$, the magnetic field direction angle $\theta$ must be greater than ${180^ \circ }$.

The angle $\beta$ between ${\mathbf{B}}$ and $\mathbf{z_1}$ of the first magnet characterizes the steering capability of ${\mathbf{B}}$. To ensure the first two springs bend coherently with the torques, $\beta$ must satisfy $\beta \in (0, 90^\circ)$. As $\beta$ approaches zero, the magnetic torque on the first magnet vanishes, leading to an extremely small bending stiffness that is impractical for fabrication. 
Conversely, an overly large $\beta$ is unfavorable for straightening the alignment part. Keeping $\beta$ small improves the directional alignment between the alignment-part magnetization and $\mathbf{B}$, thereby promoting passive straightening.
Therefore, we set $\beta  = {20^ \circ }$ so that $\theta^*_{\rm{2}}  = \gamma_{\rm{max}}+\beta={200^ \circ }$. 
$\theta^{*}_{\rm{2}}=200^\circ$ is the solution of $\theta$ to adjust the tip of SO-MAC to bend to $\gamma_{\rm{max}}$.
In summary, given parameters $\left\| {\mathbf{B}} \right\| = 40~{\rm{mT,}}~{d_{\rm{p}}} = 8~{\rm{cm,}}~\gamma_{\rm{max}} {\rm{ = 180}}^ \circ$, the ${k_{{\rm{b,1}}}},{k_{{\rm{b,2}}}}$ can be obtained with following equation 
\begin{equation}
{k_{{\rm{b}},n}} = \frac{{\left\| {\sum\limits_{i = 1}^n {{{\bf{t}}_i}} ({{\bf{z}}_i},\theta ) + {{\bf{t}}_{{\rm{eq}},n}}({\alpha _{{\rm{d}},n}}) + {{\bf{t}}_{{\rm{in}},n}}({\alpha _{{\rm{d}},n}})} \right\|}}{{{\alpha _{{\rm{d}},n}}}}.
\label{13}
\end{equation}

According to the steering pivot design, during the advancement of the SO-MAC, the steering pivot must remain stationary to assist in steering, while the alignment part is aligned with the lumen center.
The target lumen center position $\mathbf{p}_{\rm{t}}$ can be obtained from 
\begin{equation}
    {[{{\mathbf{p}}_{\rm{t}}}^{\rm{T}}~1]^{\rm{T}}} = {}^3{{\mathbf{T}}_2}(\alpha _2^*){}^2{{\mathbf{T}}_1}(\alpha _1^*){\left[ {\begin{array}{*{20}{c}}
0&0&{{d_{\rm{p}}}}&1
\end{array}} \right]^{\rm{T}}},
    \label{position}
\end{equation}
where $\alpha _1^* = \alpha _2^* = {\pi/2}$. 
The calculated $\mathbf{p}_{\rm{t}}$ remains unchanged in subsequent operations and is applicable when $n \ge 3$.

\subsubsection{$n\ge3$}
When calculating the bending stiffness of the $n$-th spring ($n \ge 3$), the designed bending angle for the $n$-th spring ${\alpha_{\mathrm{d},n}}$ is set to ${\pi/2}$ as shown in Fig. \ref{fig.motion_pattern}(b), and the number of unsupported segments $t=n$. The variables $\theta$ and ${\bm{\alpha}}\vert_{(n-1) \times 1}$, where ${\bm{\alpha}}\vert_{(n-1) \times 1}$ represents the bending angles of the first to the $(n-1)$-th segments, corresponding to the alignment of the first magnet’s axis $\mathbf{z}_1$ with the lumen center $\mathbf{p}_{\mathrm{t}}$, are obtained by solving the following optimization problem with the interior-point method \cite{byrd2000trust}:
\begin{equation}
    \begin{array}{*{20}{l}}
    {\mathop {\arg \min }\limits_{\theta ,{{\bm{\alpha }}\vert_{(n - 1) \times 1}}} \left\| {\arccos (\frac{{{{\mathbf{z}}_1} \cdot {{\mathbf{v}}_{{\rm{targ}}}}}}{{\left\| {{{\mathbf{z}}_1}} \right\|\left\| {{{\mathbf{v}}_{{\rm{targ}}}}} \right\|}})} \right\|}\\
    {s.t.\left\{ {\begin{array}{*{20}{c}}
    {0 < \theta  < {\theta_{\rm{0}} } + \delta   = {\theta _{{\rm{max}}}}}\\
    {0 < {\alpha _l} < {\alpha_2^* } + \delta  = {\alpha _{{\rm{max}}}}(l = 1,2, \ldots ,n - 1)}\\
    {\left\| {\sum\limits_{i = 1}^l {{{\mathbf{t}}_i}}  + {{\mathbf{t}}_{{\rm{eq}},l}} + {{\mathbf{t}}_{{\rm{in}},l}}} \right\| - {k_{{\rm{b}},l}}{\alpha _l} = 0}
    \end{array}} \right.}
    \end{array}
    \label{mini1}
\end{equation}
where ${{\mathbf{v}}_{{\rm{targ}}}} = \frac{{{{\mathbf{p}}_{\rm{t}}} - {\mathbf{p}}{}_1}}{{\left\| {{{\mathbf{p}}_{\rm{t}}} - {\mathbf{p}}{}_1} \right\|}}$.
The initial value of $\theta$ is ${{\theta _0} = \theta_2^* ={200^\circ }(n = 3)}$ and ${\theta _0} = \theta _{n - 1}^*\left( {n \ge 4} \right)$ where $\theta _{n - 1}^*$ is the numerical solution of $\theta$ in the last step.
In the constraints, $\delta = 10^{-4}\rm{rad}$.
And the initial value of ${{{\bm{\alpha }}\vert_{(n - 1) \times 1}}}$ is as follows
\begin{equation}
    {\bm{\alpha }^0}\vert_{(n - 1) \times 1} = \left\{ {\begin{array}{*{20}{c}}
    {{{\left[ {\begin{array}{*{20}{c}}
    {{\frac{{\alpha _1^*}}{2} }}&{{\frac{{\alpha _1^*}}{2} }}
    \end{array}} \right]}^{\rm{T}}},n = 3}\\
    {{{\left[ {\begin{array}{*{20}{c}}
    {\delta }&{{{[{\bm{\alpha }^*}\vert_{(n - 2) \times 1}]}^{\rm{T}}}}
    \end{array}} \right]}^{\rm{T}}},n \ge 4}
    \end{array}} \right.,
\end{equation}
where ${{\bm{\alpha }^*}\vert_{(n - 2) \times 1}}$ is the numerical result of ${{{\bm{\alpha }}\vert_{(n - 2) \times 1}}}$ in the last step.
For $n = 3$, $\alpha_1^*$ is directly taken from the solution obtained in the $n = 2$ case.

After solving for $\theta$ and ${\bm{\alpha}}\vert_{(n-1) \times 1}$, substituting them into (\ref{13}) enables the calculation of ${k_{{\rm{b}},n}}$, where ${\bm{\alpha}}\vert_{(n-1) \times 1}$ is used to determine $\mathbf{z}_i$ in (\ref{13}).
Iterative calculations can be performed to obtain the bending stiffness of all subsequent springs.

\section{Prototyping of SO-MAC}
\label{sec:Manufacture}
SO-MAC was fabricated with permanent magnets (NdFeB, N52, remanence ${B_r} = 1.42{\rm{T}}$, diameter ${d_{\rm{m}}} = 1.5~{\rm{mm}}$, length ${l_{\rm{m}}} = 2~{\rm{mm}}$).
The prototype has a diameter of $1.5~\rm{mm}$, a minimum tip radius of curvature of ${r_{{\rm{d,}}n}} + \frac{{{l_m}}}{2} = 2+1~{\rm{mm}}= 3~{\rm{mm}}$, and a tip bending range of 0$^\circ$ to 180$^\circ$, achieved by its first two segments (approximately 10 mm in tip length).
We set the number of segments in SO-MAC to $N=6$.
The ${k_{{\rm{b}},n}}$ can be calculated from \eqref{13} and shown in Table \ref{tab:para}.
\begin{table}[t]
    \caption{Stiffness results from simulations}
    \centering
    \begin{tabular}{c|c|c|c|c|c|c}
        \toprule
        \hline
        Spring $n$ &  $1$ &  $2$ & $3$  &  $4$ &  $5$ & $6$           \\ \hline
        \begin{tabular}[c]{@{}c@{}} ${{k_{{\rm{b}},n}\times {10^{ - 5}}}}$\\ ($   \rm{N}\cdot \rm{m}/rad$)\end{tabular} &  $3.81$ &  $13.37$ & $18.71$  &  $21.28$ &  $22.54$ & $23.17$  \\
        \hline
        \bottomrule
    \end{tabular}
    \label{tab:para}
\end{table}

\begin{figure}[t]
  \centering
  \includegraphics[ width=0.3\textwidth]{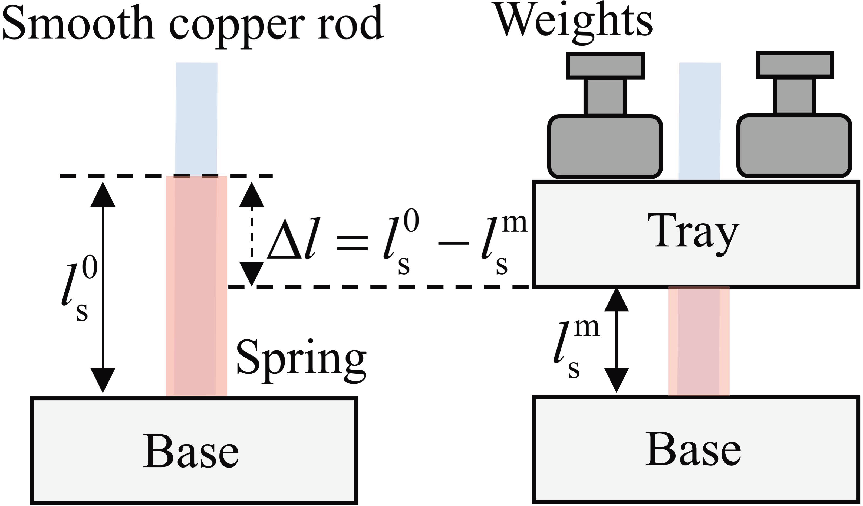}
  \caption{Measurement principle of compression stiffness. 
 The copper rod is firmly connected to the base.
The rod and the tray have a clearance fit, ensuring smooth sliding with negligible friction and no wobbling.
Both the base and the tray were fabricated using photocuring 3D printing (Wenext Co., Ltd., China).
 All lengths are measured using a vernier caliper (minimum resolution: 0.01 mm, Deli, China).
 }
  \label{fig.stiff_measure}
\end{figure}

According to \cite{wahl1944mechanical}, the relationship between ${{k_{{\rm{b}},n}}}$ and ${{k_{{\rm{c}},n}}}$ as 
\begin{equation}
    {k_{{\rm{b,}}n}} = \frac{{Ed_n^4}}{{32c{D_n}}}\left( {\frac{1}{{1 + \frac{E}{{2G}}}}} \right) =  \frac{{ED_n^2}}{{2\left( {2G + E} \right)}} ~ {k_{{\rm{c,}}n}}
    \label{kb}
\end{equation}
where $c$ is the number of active coils, $d_n$ is the diameter of the wire of spring $n$, $D_n$ is the nominal diameter of spring $n$, $E$ and $G$ are the elastic modulus and shear modulus of the spring.
Since the measurement of  ${{k_{{\rm{b}},n}}}$ is difficult with such a short axial length $l_{\rm{s}}$, we can calculate  ${{k_{{\rm{b}},n}}}$ by \eqref{kb} after measuring ${{k_{{\rm{c}},n}}}$.
We apply an axial force to the springs by using the combined gravity of the weights and tray with a total mass of $m$.
The spring is placed to a smooth copper rod in Fig. \ref{fig.stiff_measure}. 
By conducting five repeated trials, we measure the resulting average change in the spring length, denoted as $\Delta \bar l$.
Then we can calculated ${k_{{\rm{c}},n}} = \frac{{mg}}{{\Delta \bar l}}$.

We ordered a batch of stainless steel springs (${l_{\rm{s}}} = 3.14~\rm{mm}$, $E=200\rm{GPa}$, $G=E/(2+2v)$ and $v=0.3$ is Poisson's ratio, Shenzhen Haoxin United Industrial Co., Ltd.) and selected the springs that closely match the design values in Table \ref{tab:para} to manufacture the SO-MAC.
Using (\ref{kb}), we select wire diameters $d_n$ of 0.1mm and 0.15mm and original length of the springs $l_{\rm{s}}^0=\pi \approx 3.14 ~\rm{mm}$. 
The design specification involved varying the outer diameter $D=D_n+d_n$ of the springs within the range of $1\rm{mm}$ to $1.5\rm{mm}$.
In addition, the active coil number $c$ ranged from $3$ to $7$ coils, providing a wide range of spring selection options from ${k_{{\rm{b}},n}}=2.77 \sim 53.95\times {10^{ - 5}}{\rm{N \cdot m/rad}}$.

\begin{figure}[t]
  \centering
  \includegraphics[ width=0.3\textwidth]{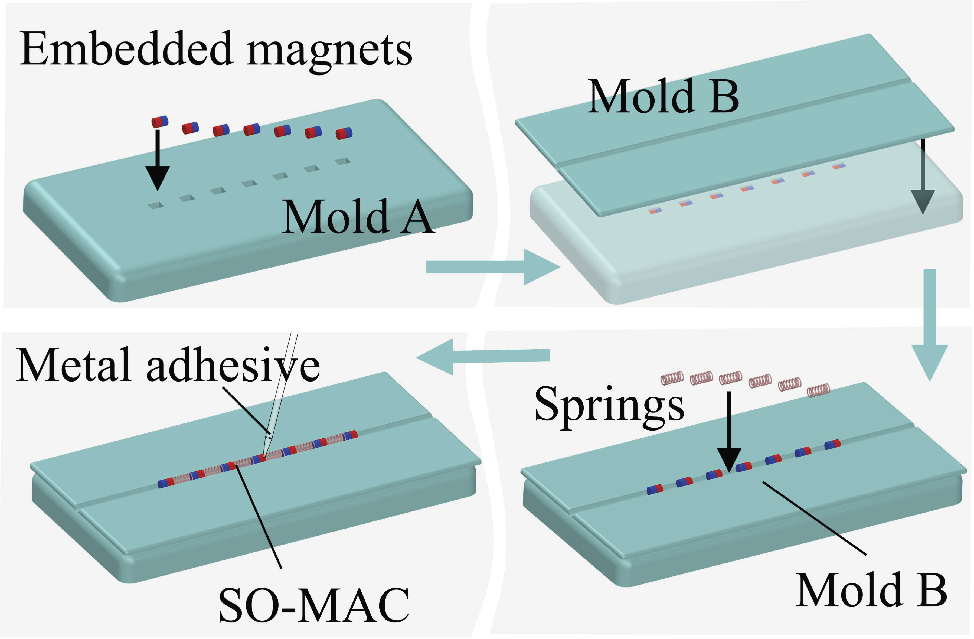}
  \caption{Prototyping of SO-MAC. The attraction from the positioning magnets in mold A and the constraint of the groove on mold B ensure precise spring-to-magnet attachment without relative motion during manufacturing.}
  \label{fig.manufacture}
\end{figure}

The bending stiffnesses ${k_{{\rm{b}},n}^*}$ of the selected springs are shown in Table \ref{tab:selectedSprings}, where ${e_{{\rm{kb,}}n}} = \left| {k_{{\rm{b}},n}^* - {k_{{\rm{b}},n}}} \right|/{k_{{\rm{b}},n}}$ represents the relative errors between the actual measured bending stiffnesses and the designed values.
We 3D‑printed the molds (Fig.~\ref{fig.manufacture}), placed magnets at spacing $l_{\rm{s}}$ in mold A, and added cover mold B. 
The SO‑MAC magnets were fixed in mold B by attraction to the magnets in mold A. 
The springs were then inserted and bonded using metal adhesive (Deli, China).

\begin{table}[t]
    \caption{Parameters of the selected springs}
    \centering
    \setlength{\tabcolsep}{1.5mm}{
    \begin{tabular}{c|c|c|c|c|c|c}
        \toprule
        \hline
        Spring $n$ &  $1$ &  $2$ & $3$  &  $4$ &  $5$ & $6$           \\ \hline
        \begin{tabular}[c]{@{}c@{}} ${{k_{{\rm{b}},n}^* \times {10^{ - 5}}}}$\\ ($   \rm{N}\cdot \rm{m}/rad$)\end{tabular} &  $3.96$ &  $13.69$ & $18.24$  &  $20.94$ &  $21.54$ & $23.83$  \\
        \hline
        \begin{tabular}[c]{@{}c@{}} ${{k_{{\rm{c}},n}^*}}$\\ ($   \rm{N}/ \rm{m}$)\end{tabular} &  $97.48$ &  $265.93$ & $419.77$  &  $489.73$ &  $495.63$ & $548.49$  \\
        \hline
        \begin{tabular}[c]{@{}c@{}} ${{e_{{\rm{kb},n}}}}$\\ ($\%$)\end{tabular} &  $3.94$ &  $2.39$ & $2.51$  &  $1.60$ &  $4.44$ & $2.85$  \\
        \hline
        \bottomrule
    \end{tabular}}
    \label{tab:selectedSprings}
\end{table}

\section{Sensitivity Analysis of Pivot Performance}
\label{FollowEvaluation}
In the stiffness design of the SO-MAC, the aim is to maintain the steering pivot at the steering part with the advancing direction of $\gamma = 180^\circ$. 
Here, $\gamma$ is also referred to as the target lumen direction in this paper.
To evaluate the robustness, we investigated whether the SO-MAC could maintain its steering pivot under variations in $\gamma$, the magnetic field strength of the external magnetic field, and the lumen center distance $d_{\rm{p}}$.

\subsection{Shapes of SO-MAC During Advancement Toward $\gamma$}
To obtain SO-MAC shapes during motion toward $\gamma$, we solve (\ref{gamma}) for ${{\bm{\alpha }}\vert_{t \times 1}}$ and the field direction $\theta$ given $\gamma$ and the bending stiffnesses, using (\ref{normEQ}) as the third equilibrium constraint; the initial aligned state (tip aligned with the lumen center along $\gamma$) is ${{\bm{\alpha^* }}\vert_{2 \times 1}} = {\left[ {\begin{array}{*{20}{c}}
{{\alpha^* _1}}&{{\alpha^* _2}}
\end{array}} \right]^{\rm{T}}}$ and $\theta$.
\begin{equation}
\left\{ {\begin{array}{*{20}{c}}
{{\alpha _1} + {\alpha _2} - \gamma  = 0,t = 2}\\
{{{\left\| {{s_{\rm{a}}} \  \arccos (\frac{{{{\mathbf{z}}_1} \cdot {{\mathbf{v}}_{{\rm{targ}}}}}}{{\left\| {{{\mathbf{z}}_1}} \right\|\left\| {{{\mathbf{v}}_{{\rm{targ}}}}} \right\|}})} \right\|}^2} = 0,t \ge 3}\\
{{s_n} \ \left( {\left\| {\sum\limits_{i = 1}^n {{{\mathbf{t}}_i}}  + {{\mathbf{t}}_{{\rm{eq}},n}} + {{\mathbf{t}}_{{\rm{in}},n}}} \right\| - {k_{{\rm{b}},n}}{\alpha _n}} \right) = 0,n = 1, \cdots ,t}
\end{array}} \right.
    \label{gamma}
\end{equation}
A scaling factor of ${s_{\rm a}=10}$ is introduced in the second equilibrium constraint to enforce alignment between the SO‑MAC alignment part and the lumen center, consistent with (\ref{mini1}).
\begin{figure}[t]
  \centering
  \includegraphics[ width=0.45\textwidth]{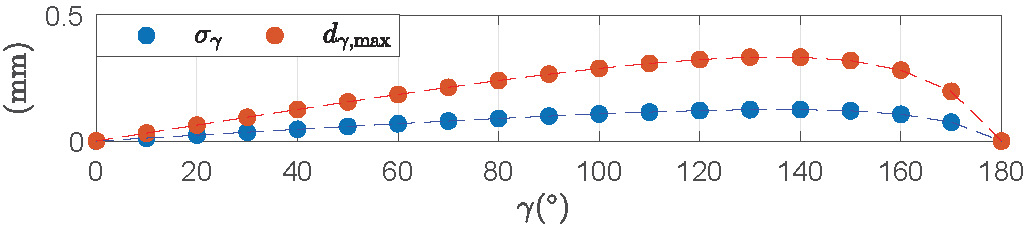}
  \caption{Theoretical steering pivot errors under different target lumen direction $\gamma$. The calculation is done under the designed parameters ($\left\| {\mathbf{B}} \right\| = 40~{\rm{mT,}}~{d_{\rm{p}}} = 8~{\rm{cm}}~$).}
  \label{fig.FLerror}
\end{figure}

We set the initial values as follows
\begin{equation}
\left\{ {\begin{array}{*{20}{c}}
{\left\{ {{\theta _{0,t}} = \gamma  + \beta (\frac{\gamma }{\pi }),{\mathbf{\alpha }^0}\vert_{t \times 1} = {{\left[ {\begin{array}{*{20}{c}}
{\frac{\gamma }{2}}&{\frac{\gamma }{2}}
\end{array}} \right]}^{\rm{T}}}} \right\},t = 2}\\
{\left\{ {{\theta _{0,t}} = \theta _{t - 1}^*,{\mathbf{\alpha }^0}\vert_{t \times 1} = {{\left[ {{{\begin{array}{*{20}{c}}
\delta &{\left[ {{\mathbf{\alpha }^*}\vert_{(t - 1) \times 1}} \right]}
\end{array}}^{\rm{T}}}} \right]}^{\rm{T}}}} \right\},t \ge 3}
\end{array}} \right.
\label{initialConditions}
\end{equation}

Substituting $\alpha_1^*$ and $\alpha_2^*$, which are obtained from the numerical results of $t=2$ in (\ref{gamma}), into (\ref{position}), we obtain the lumen center position ${\mathbf{p}}_{\rm{t}}$, which remains unchanged in the subsequent calculations of advancement towards the given $\gamma$.
For $t \geq 3$, the subsequent shapes ${{\bm{\alpha }}\vert_{t \times 1}}$ of the SO-MAC and the corresponding external magnetic field directions $\theta$ can be calculated using (\ref{gamma}).

\subsection{Steering Pivot Performance under Varying $\gamma$}
To quantify the steering pivot ability at different $\gamma$, we utilized $\gamma  ={0^ \circ }, {10^ \circ },{20^ \circ }, {\cdots},{180^ \circ }$ but with constant $\left\| {\mathbf{B}} \right\| = 40~{\rm{mT}}$ and $d_p=8~\rm{cm}$ for simulation. 
During the SO-MAC advancement, the number of movable segments in the SO-MAC, denoted by $t$, varies from $2$ to $N=6$ segments. 
In each state, the position of the steering pivot magnet is represented as ${\mathbf{p}}_t^{(\gamma )}$, which can be calculated from (\ref{position_allmag}).
We then calculated the root mean square error (RMSE) ${\sigma _\gamma}$ and maximum variation ${{d_{\gamma,\max }}}$ of the positions of the steering pivot magnet during the advancement of the SO-MAC as 
\begin{equation}
    {\sigma _\gamma } = \sqrt {\frac{1}{{N - 1}}{{\sum\limits_{t = 2}^N {\left( {\left\| {{\mathbf{p}}_t^{(\gamma )} - \overline {{\mathbf{p}}^{(\gamma )}} } \right\|} \right)} }^2}} 
    \label{sigma_error}
\end{equation}
and
\begin{equation}
    {d_{\gamma,\max }} = \max \{ \left\| {{\mathbf{p}}_i^{(\gamma )} - {\mathbf{p}}_j^{(\gamma )}} \right\|\} ,\forall i,j \in \left\{ {2,3, \ldots ,N} \right\} \wedge i \ne j.
    \label{max_error}
\end{equation}
${\overline {{\mathbf{p}}^{(\gamma )}} }$ refers to the average position of all ${\mathbf{p}}_t^{(\gamma )}$ positions along the target lumen direction $\gamma$.

 \begin{figure*}[t]
 \centering
  \includegraphics[ width=0.95\textwidth]{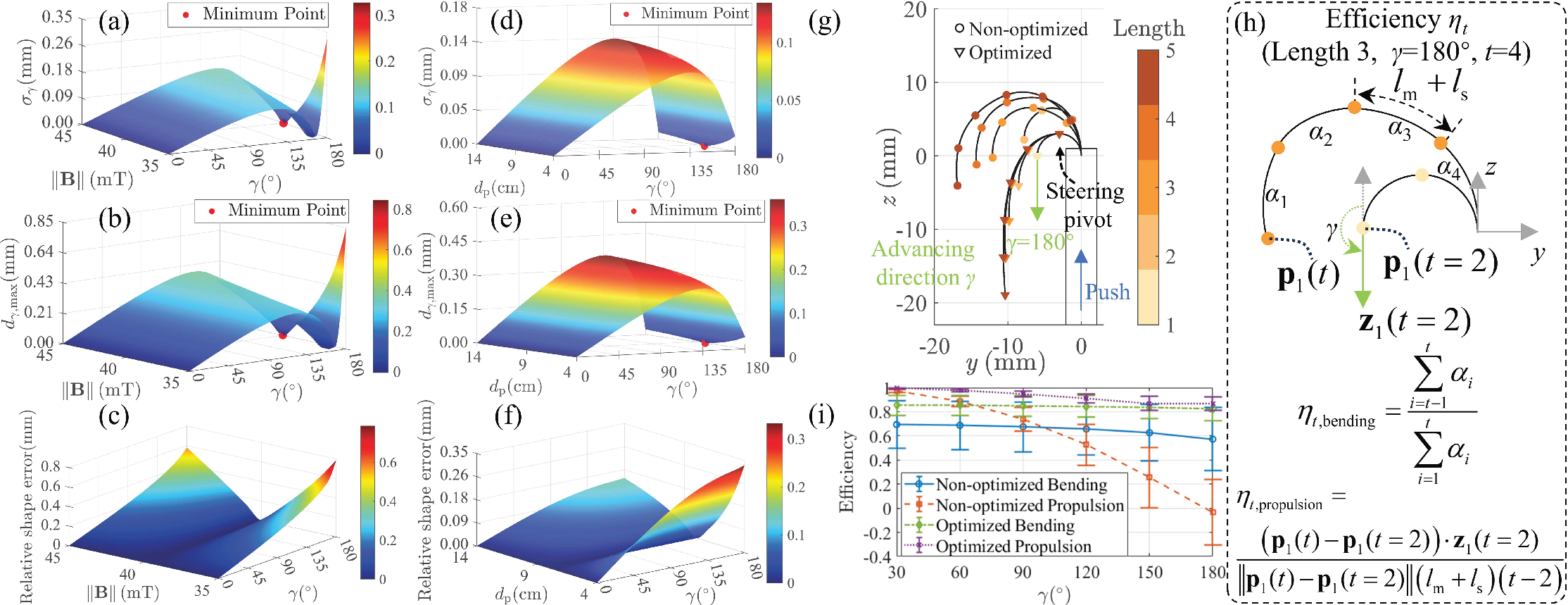}
    \caption{
    The effects of variations in magnetic field strength $\left\| {\mathbf{B}} \right\|$ and lumen center distance $d_{\rm{p}}$ on the steering pivot errors $\sigma_\gamma$ and ${d_{\gamma,\max}}$, as well as the relative shape error compared to the standard shape under the designed parameters, are examined. The results demonstrate that the steering pivot of the SO-MAC remains nearly unaffected within certain ranges of these parameter variations.
(a)–(c) illustrate the impact of varying $\left\| {\mathbf{B}} \right\|$ with a fixed ${d_{\rm{p}}}=8~\rm{cm}$, while (d)–(f) show the effect of varying ${d_{\rm{p}}}$ with a fixed $\left\| {\mathbf{B}} \right\|=40~\rm{mT}$.
(g)–(i) evaluate the bending and propulsion efficiencies of the SO‑MAC, showing that an extreme non‑optimized gradient stiffness causes dispersed bending and reduced propulsion.
    }
    \label{fig.steering pivot_evaluation_multiPara}
\end{figure*}

From the results in Fig. \ref{fig.FLerror}, we can see that the position changes of the steering pivot magnet are very small, with a maximum value of $0.13\rm{mm}$ for ${\sigma _\gamma}$ and a maximum value of $0.33\rm{mm}$ for ${{d_{\gamma,\max }}}$.   
Dividing ${d_{\gamma ,\max }},{\sigma _\gamma }$ by the total length of the SO-MAC tip, which is $32.84~\rm{mm}$, we obtained mean relative errors ${e_r}\left( {{\sigma _\gamma }} \right),{e_r}\left( {{d_{\gamma ,\max }}} \right)$ with a maximum of $0.38\% ,1.02\%$.
Therefore, during the advancement it is reasonable to conclude that SO-MAC also maintains a steering pivot under different bending angles $\gamma \in \left[ {{0}^\circ,{{180}^ \circ }} \right]$.
\subsection{Steering Pivot Performance under Varying Magnetic Field Strength and Lumen Center Distance}
To validate the steering pivot performance under varying magnetic field strength $\left\| {\mathbf{B}} \right\|$ and lumen center distance ${d_{\rm{p}}}$, we calculated the distributions of $\sigma _\gamma$ and ${d_{\gamma,\max }}$, as shown in Fig. \ref{fig.steering pivot_evaluation_multiPara}. 
$\left\| {\mathbf{B}} \right\|$ was varied from 35 mT to 45 mT (with a sampling interval of 1 mT), and ${d_{\rm{p}}}$ was varied from 4 cm to 14 cm (with a sampling interval of 1 cm) and $\gamma$ was varied in the same way as Fig. \ref{fig.FLerror}. 
In Fig. \ref{fig.steering pivot_evaluation_multiPara}(a), (b), (d), and (e), the marked minimum point represents the state with zero steering pivot errors under the designed parameters. 
The simulation results demonstrate that the $\sigma_\gamma$ of the steering pivot remains below 0.35 mm and 0.14 mm for the variations of $\left\| {\mathbf{B}} \right\|$ and ${d_{\rm{p}}}$, respectively. 
The ${d_{\gamma,\max }}$ remains below 0.85 mm and 0.4 mm, respectively.
From Fig. \ref{fig.steering pivot_evaluation_multiPara}(d) and (e), it can be observed that variations in ${d_{\rm{p}}}$ have a minimal impact on the steering pivot errors.
This suggests that SO-MAC can reliably maintain a stable steering pivot within the simulated parameter ranges.

Furthermore, the shape stability of the SO-MAC was verified under varying $\left\| {\mathbf{B}} \right\|$ and ${d_{\rm{p}}}$, as shown in Fig. \ref{fig.steering pivot_evaluation_multiPara}(c) and (f). The standard shapes, defined as the SO-MAC’s shapes under the designed conditions ($\left\| {\mathbf{B}} \right\| = 40~{\rm{mT,}}{d_{\rm{p}}} = 8~{\rm{cm}}$) to move in the steering-pivot manner towards lumen direction $\gamma$, were used as references. 
At a given steering direction $\gamma$, field magnitude $\lVert \mathbf{B} \rVert$, and pivot distance $d_{\rm p}$, $e_{\gamma,\lVert \mathbf{B} \rVert,d_{\rm p}}$ denotes the relative shape error during steering-pivot-based advancement. Following the RMSE definition in (\ref{sigma_error}), we compute the RMSE between the magnet-position vector of the standard (reference) shape and that of the simulated shape at each of five advancement lengths along the same $\gamma$ direction, and define $e_{\gamma,\lVert \mathbf{B} \rVert,d_{\rm p}}$ as the mean RMSE over the five lengths.
Fig. \ref{fig.steering pivot_evaluation_multiPara}(c) demonstrates that the shape variation due to $\left\| {\mathbf{B}} \right\|$ does not exceed 0.8 mm, while Fig. \ref{fig.steering pivot_evaluation_multiPara}(f) shows that the shape variation due to ${d_{\rm{p}}}$ is limited to a maximum of 0.35 mm. 
These results indicate that under varying $\left\| {\mathbf{B}} \right\|$ and ${d_{\rm{p}}}$, the SO-MAC is highly consistent with the standard shape, ensuring robust shape stability. 
This highlights the strong dynamic range of the SO-MAC, allowing the system to remain stable under dynamic changes in external conditions.

\subsection{Bending and Propulsion Efficiency}
A gradual increase in stiffness is essential for interventional catheters \cite{lawrence2005materials,yang2025degradable,dreyfus2024dexterous}, yet excessive stiffening has its limit. Beyond this limit, non‑optimized stiffness can reduce propulsion efficiency—hindering effective forward displacement—and lead to dispersed rather than concentrated bending.
Fig. \ref{fig.steering pivot_evaluation_multiPara}(g) compares the motion of the optimized‑stiffness SO‑MAC with that of a non‑optimized configuration ${\left[ {{k_{{\rm{b,1}}}},{k_{{\rm{b,2}}}},2{k_{{\rm{b,6}}}},2{k_{{\rm{b,6}}}},4{k_{{\rm{b,6}}}},4{k_{{\rm{b,6}}}}} \right]^{\rm{T}}}$. 
For all lengths, the first magnet was aligned toward the same lumen center at $\gamma  = {180^ \circ }$. In the non‑optimized case, the first two stiffness values were identical to those of the SO‑MAC to ensure identical initial postures. The optimized SO‑MAC maintained a fixed steering‑pivot position, concentrated bending, and effective advancement along $\gamma  = {180^ \circ }$, while the non‑optimized version showed regression or stagnation.

To quantify bending and propulsion efficiency, we defined bending efficiency (Fig. \ref{fig.steering pivot_evaluation_multiPara}(h)) as the ratio of the bending angle of the steering part (the two proximal segments) to the total bending angle of all segments at a given $\gamma$. The mean and standard deviation of bending efficiency for five successive lengths were plotted in Fig. \ref{fig.steering pivot_evaluation_multiPara}(i). 
Starting from length 1, four propulsion cycles extended the SO‑MAC to length 5. 
The propulsion efficiency (Fig. \ref{fig.steering pivot_evaluation_multiPara}(h)) was defined as the ratio between the projection of the tip displacement along the advancing direction and the total push distance. 
Its average over four cycles was also shown in Fig. \ref{fig.steering pivot_evaluation_multiPara}(i).

While the non‑optimized design performed comparably to the optimized one at small advancing angles, the optimized stiffness maintained both bending and propulsion efficiencies above 0.8 across $\gamma  = {30^ \circ } - {180^ \circ }$ with small, consistent variances. 
In contrast, the non‑optimized configuration exhibited sharply reduced efficiencies and markedly increased variance as the advancing angle $\gamma$ grew. 
These results demonstrate that the optimized stiffness‑increase scheme ensures robust and stable performance across directions, thereby revealing the physical limit of stiffness augmentation.

\section{Experiments and Results}
\label{sec:exp}
\subsection{System Setup}
\label{System Setup}
\begin{figure}[t]
  \centering
  \includegraphics[ width=0.49\textwidth]{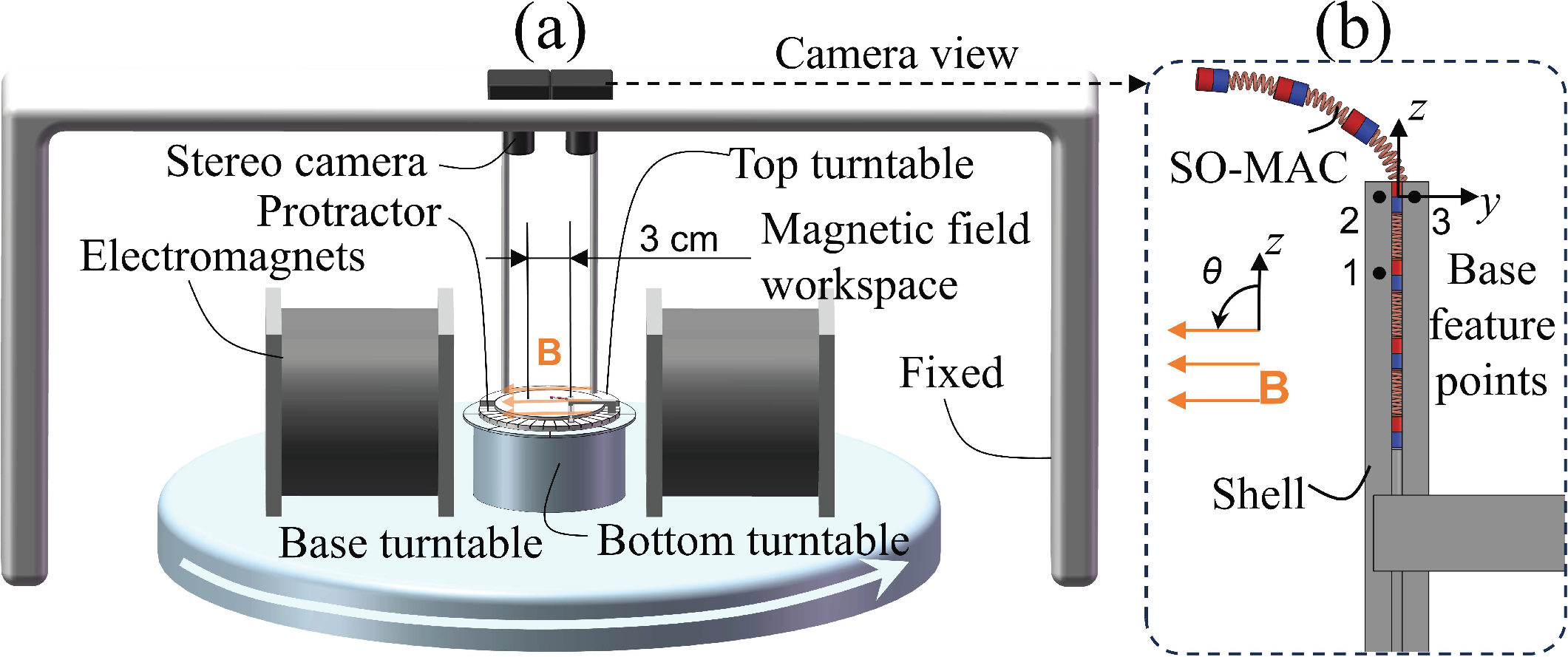}
  \caption{Diagram of experimental setups for experiments \ref{Kinetostatic Model Validation} and \ref{steering pivot Validation}. 
  (a) Two electromagnets, the bottom turntable and the base turntable are rigidly connected as a single unit, while the platform on which the SO-MAC operates is rigidly connected to the stereo camera. By rotating the base turntable and reading the values on the protractor, the input magnetic field direction $\theta$ can be controlled.
      (b) SO-MAC is partially fixed within a shell, with the remaining part free, allowing its free length to be controlled via a pushrod. 
      Base feature points assist the stereo camera in reconstructing the SO-MAC's world frame for comparison with theoretical results.}
  \label{fig.equipments}
\end{figure}

Fig.~\ref{fig.equipments}(a) shows the setup for the first two experiments: two DC power supplies (IT6942A, ITECH) independently drive the two electromagnetic coils, and a calibrated stereo camera (H200S, JIERUIWEITONG; $1920\times1080$ each; mean reprojection error 0.0916 pixels) mounted on the top turntable records the SO-MAC 3D shape. 
A protractor is fixed on the top turntable, while the coils, bottom turntable, and base turntable form a rigid assembly; $\theta$ is set by rotating the base turntable and adjusting the top-bottom turntable angle. 
Experiments were performed with the SO-MAC in an $\sim$3 cm workspace between the electromagnets, where the field was 38--42 mT (median 40 mT; Tesla meter HT20, Hengtong Magneto-electric, Shanghai).
Because the two-electromagnet field is nonuniform over the 3 cm workspace (two 1.5 cm regions with opposite gradients, mean $\approx 266.67\,\mathrm{mT/m}$, 42$\xrightarrow{}$38$\xrightarrow{}$42 mT), the gradient-induced extra torque on the steering pivot spring is an order of magnitude smaller than the nominal magnetic torque, the pivot’s mean fluctuation is only 0.22 mm (assessed along the $\gamma$ from 10$^\circ$ to 180$^\circ$ at 10$^\circ$ intervals), and simulations in Section \ref{FollowEvaluation} show negligible effects on SO-MAC’s shape deformation, supporting the validity of this setup for model validation.

In the third experiment (Section \ref{Navigation}), an external permanent magnet was mounted on a robotic manipulator (UR5e, Universal Robots), as shown in Fig. \ref{fig.exp3_procedure}.
The propulsion device, a linear actuator (FSK30, Chengdu Fuyu), is for tether actuation. 
The RGB camera (Diameter 2 mm, Length 5 mm, $400\times 400$ pixels, OV6946, ANYVIEW, China) is rigidly attached to the first magnet of the SO-MAC. 
The SO‑MAC equipped with an RGB camera and a NiTi‑wire tether is shown in the attached video.

\subsection{Kinetostatic Model Validation}
\label{Kinetostatic Model Validation}
We validated the proposed kinetostatic model by comparing predicted and measured SO-MAC shapes under open-loop field inputs $\theta\in\{30^\circ,60^\circ,90^\circ,120^\circ,150^\circ,180^\circ\}$ and varying lengths. 
The embedded-magnet centers served as shape feature points; the $o$-$xyz$ world frame origin was set at the midpoint of base feature points 2 and 3 (Fig.~\ref{fig.equipments}(b)), i.e., the base-magnet position. 
For each $\theta$, five lengths were recorded; from each stereo image pair, we extracted the pixel coordinates of the three base feature points and magnet centers and reconstructed their 3D coordinates by triangulation, treating the SO-MAC and base points as a rigid body.
Because the SO-MAC shape is approximately planar, magnet positions were projected onto the plane defined by the three base feature points; using the Length~1 base points as the reference, Lengths~2--5 were rigidly registered to obtain rotation/translation and align all shapes in a common frame, yielding an average registration error of 0.039 mm, computed as the root mean square error (RMSE) between the registered and reference base points.

\begin{figure}[t]
  \centering
  \includegraphics[ width=0.485\textwidth]{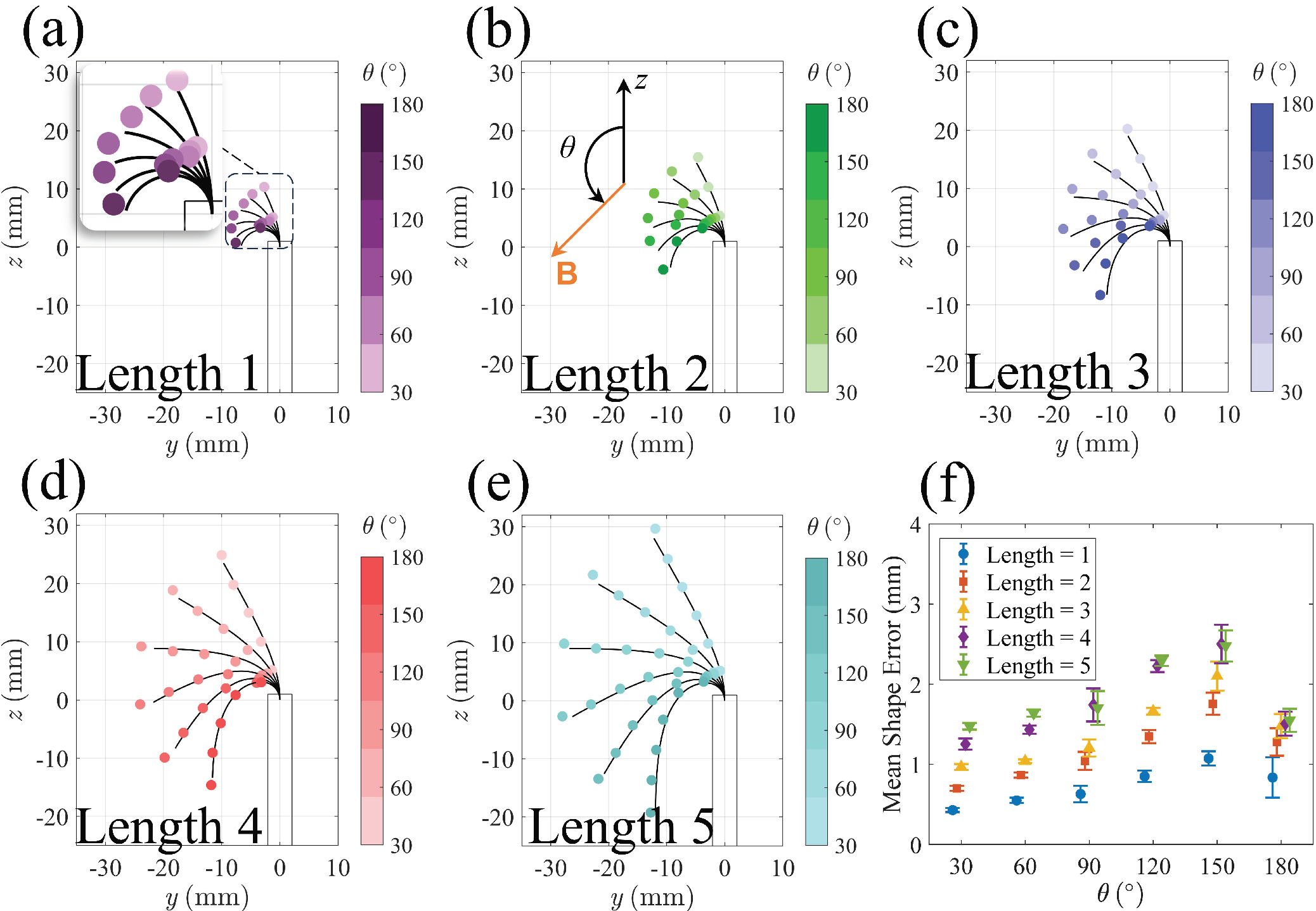}
  \caption{(a)-(e) Measured data and theoretical shapes of SO-MAC from the kinetostatic model simulation at different magnetic field direction $\theta$.
			The markers represent the measured positions of the magnets on SO-MAC.
            The black curves represent the theoretical shapes corresponding to each SO-MAC configuration.
			We use different colors to distinguish the data for different values of $\theta$. 
            (f) Distribution of the mean shape errors of SO-MAC and their corresponding standard deviations across five repeated measurements. 
            Compared with the theoretical value of the model, the overall mean shape error of SO-MAC is 1.39 mm.
            }
  \label{fig.exp1_AD}
\end{figure}
The simulated and experimental shapes of the SO-MAC under different magnetic field directions $\theta$ are shown in Fig. \ref{fig.exp1_AD}(a)–(e). 
Each subfigure corresponds to a fixed number of unsupported segments $t$ and includes six SO-MAC shapes under six distinct $\theta$ values. 
The subfigures are labeled as Length $t$–1.

The shape error $e_{t,\mathrm{Shape}}$ is computed analogously to (\ref{sigma_error}) as the RMSE between the measured and theoretical SO-MAC shapes for $t$ movable magnet--spring segments, evaluated over the $t{+}1$ magnet positions.

For each magnetic field angle $\theta$ and SO-MAC length, five repeated measurements were conducted to evaluate the shape errors. 
The distribution of the mean shape errors and their corresponding standard deviations is shown in Fig. \ref{fig.exp1_AD}(f).
As the SO-MAC length increases (i.e., from Length 1 to Length 5), the mean shape error exhibits a rising trend, primarily due to the accumulation of manufacturing errors.
Across all measurements, the maximum standard deviation is only 0.25 mm, which is significantly smaller than the SO-MAC's diameter (1.5 mm), indicating high repeatability and shape consistency under identical magnetic fields.
The overall mean shape error across all 150 shape samples is ${\bar e_{t,\mathrm{Shape}}} = 1.39 \pm 0.56~\mathrm{mm}$, which remains small relative to the total SO-MAC length of 32.84 mm. 
These results validate the accuracy of the proposed kinetostatic model.

\subsection{Steering Pivot Validation}
\label{steering pivot Validation}
The steering pivot experiment was conducted using the same setup shown in Fig. \ref{fig.equipments} to validate the accuracy of SO-MAC advancement under the steering-pivot strategy proposed in Section \ref{FollowEvaluation}.
Using the theoretical magnetic field directions $\theta$ derived in Section \ref{FollowEvaluation}, we performed open-loop control to guide the SO-MAC along the desired advancing direction $\gamma$. The SO-MAC shapes during advancement were recorded following the methodology described in Section \ref{Kinetostatic Model Validation}.
In this context, $\theta$ represents the predicted magnetic field direction that aligns the SO-MAC with the target lumen center. The results demonstrate that the SO-MAC can achieve effective displacement along $\gamma$ while maintaining steering pivot.

\begin{figure}[t]
  \centering
  \includegraphics[ width=0.49\textwidth]{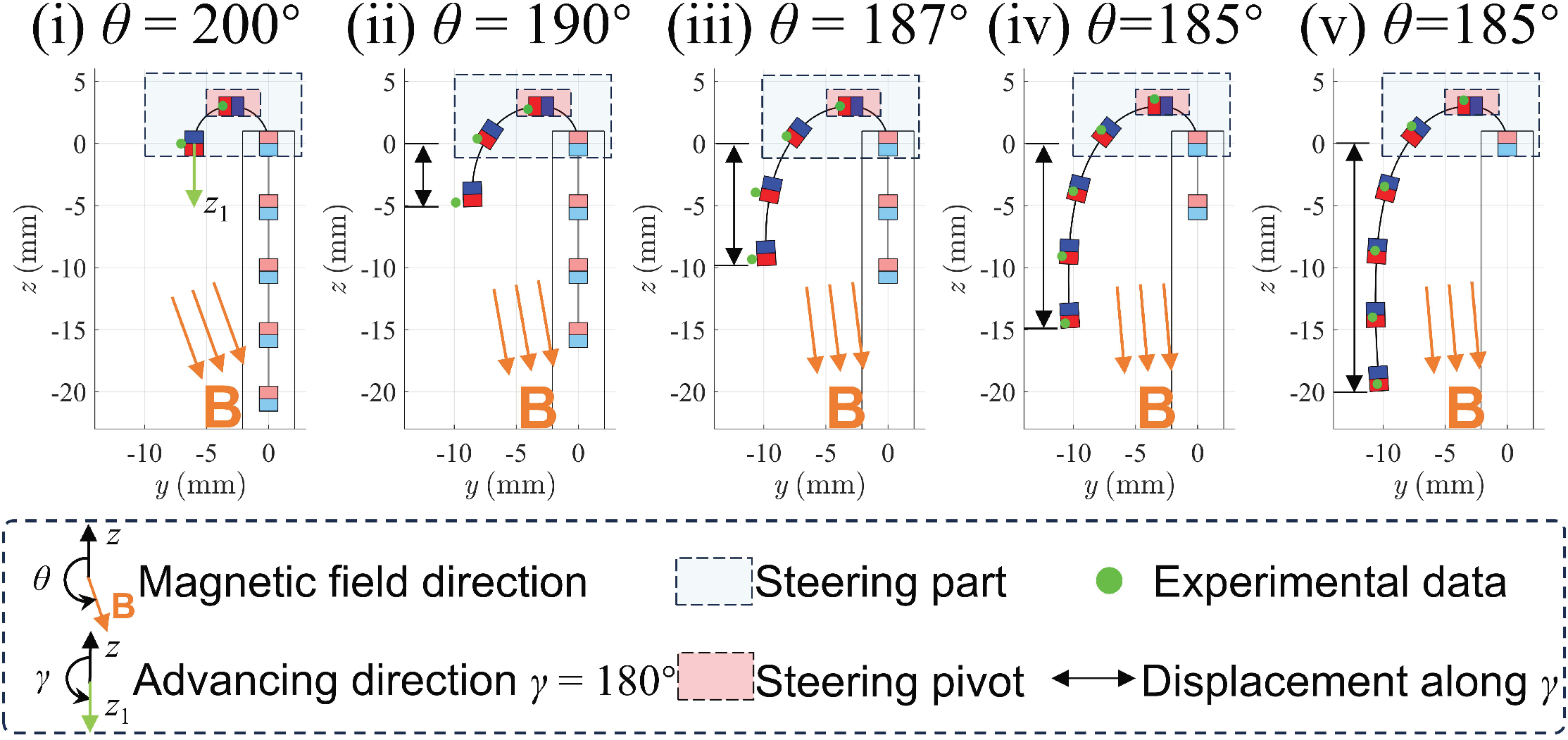}
  \caption{The process of SO-MAC advancement along the target lumen direction of $\gamma=180^{\circ}$ with steering pivot manner.
  $\theta$ represents the control input corresponding to different unsupported segment numbers.
  We can clearly observe that the SO-MAC ensures the steering pivot within the steering part remains stationary while generating effective displacement along the desired $\gamma=180^\circ$ direction.
}
  \label{fig.exp2_1}
\end{figure}

Fig. \ref{fig.exp2_1} illustrates a representative case of the most challenging scenario, where the SO-MAC advances along the $\gamma = 180^\circ$ direction.
The figure shows that the position of the steering pivot magnet remains consistently fixed throughout the advancement, effectively stabilizing the steering parts of the SO-MAC near a specific location.
These observations confirm that the steering-pivot strategy functions as intended under $\gamma = 180^\circ$.
Furthermore, the displacement of the alignment part follows the $\gamma = 180^\circ$ direction, closely matching the theoretical design to produce effective tip displacement.
When navigating through narrow orifices with sharp turning angles, the SO-MAC is expected to exhibit strong steering capabilities using a short tip to achieve large deflections. As shown in Fig. \ref{fig.exp2_1}(i), the first two segments of the SO-MAC (approximately 10 mm in length) can be bent close to $180^\circ$ by adjusting the direction of the external magnetic field $\theta$, demonstrating its maximum tip bending capability.

\begin{figure}[t]
  \centering
  \includegraphics[ width=0.35\textwidth]{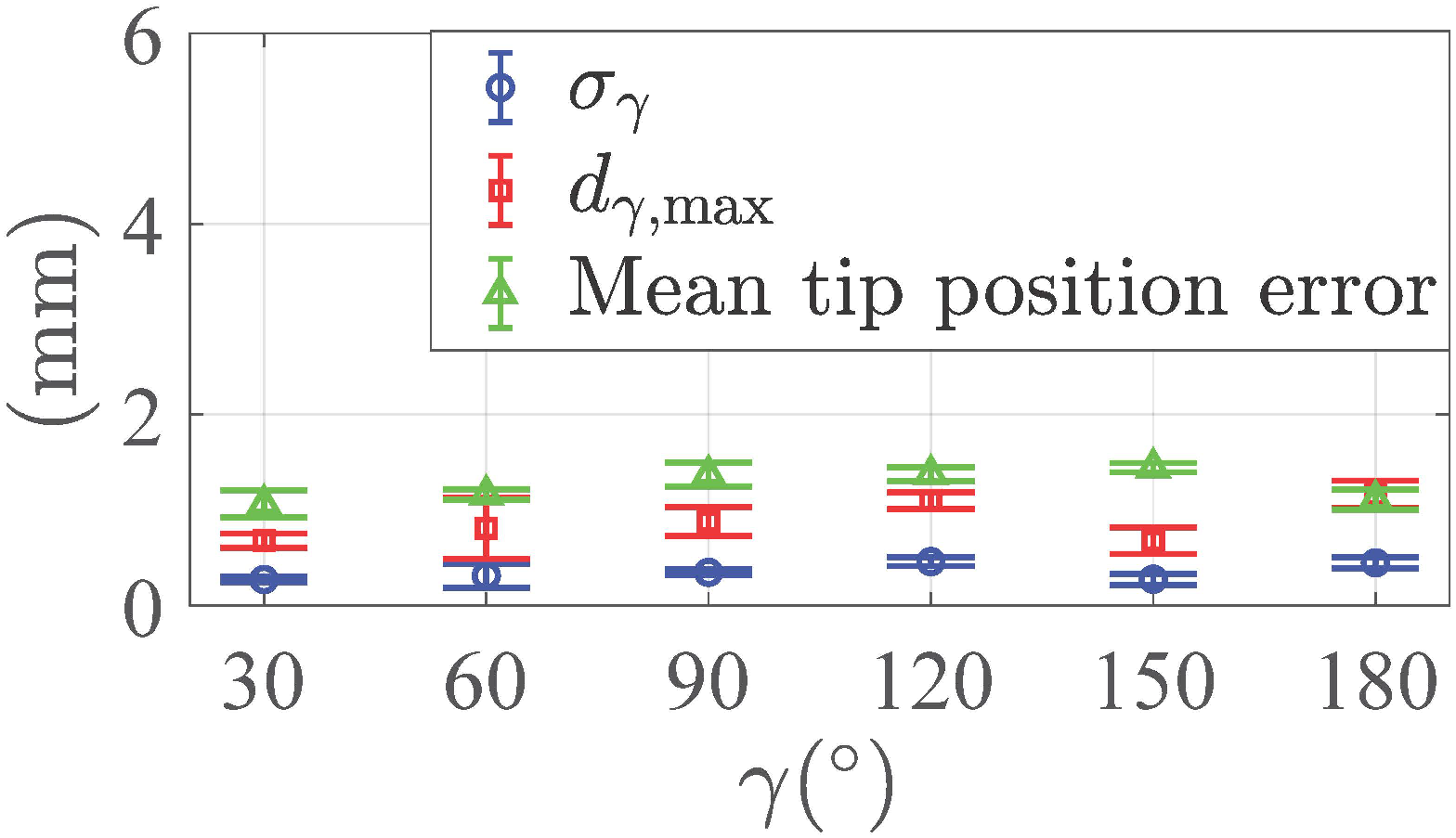}
  \caption{Steering pivot errors of SO-MAC at different $\gamma$.
Each point represents five independent repeated experiments conducted in the current $\gamma$ direction, with the central point indicating the mean steering pivot error (red squares for ${d_{\gamma ,\max }}$ and blue circles for $\sigma_\gamma$). The error bars denote the standard deviations.
The green triangles represent the mean error between the tip position and the theoretical position during the SO-MAC's motion along $\gamma$.
}
  \label{fig.flerror_real}
\end{figure}

Similar to the open-loop experiment at $\gamma=$ 180$^\circ$, we also conducted experiments at $\gamma= $ 30$^\circ$, 60$^\circ$, 90$^\circ$, 120$^\circ$, 150$^\circ$.
Each steering pivot motion along each $\gamma$ direction was independently repeated five times.
Similarly, in this experiment, the measured SO-MAC shapes were ensured to be in the same frame by registering the three base feature points in Fig.\ref{fig.equipments}(b). 
The mean registration error of the base feature points in this experiment was 0.049 mm.
We evaluated the actual errors of the steering pivot based on criteria (\ref{sigma_error}) and (\ref{max_error}), as shown in Fig. \ref{fig.flerror_real}. 
Using the data from all steering pivot experiments, we calculated the mean values of ${d_{\gamma ,\max }}$ and $\sigma_\gamma$, which were ${{{\bar d}_{\gamma ,\max }} = 0.89 \pm 0.25{\rm{mm}}}$ and ${{{\bar \sigma }_\gamma } = 0.35 \pm 0.10{\rm{mm}}}$.
Both standard deviations (0.25 mm and 0.10 mm) were significantly smaller than the diameter of the SO-MAC (1.5 mm). 
A slightly larger standard deviations at $\gamma = 60^\circ$ (0.32 mm and 0.12 mm) likely resulted from small variations in the identified magnet center during stereo‑vision triangulation, yet this localization uncertainty remained far below the magnet size.
Fig. \ref{fig.flerror_real} presents the mean tip position errors of the SO-MAC during advancement along various $\gamma$ directions. 
The overall tip position error was $1.26 \pm 0.17~\mathrm{mm}$.
These results show that the proposed SO-MAC steers effectively along the desired advancing direction while maintaining a fixed steering pivot for concentrated bending.

\subsection{Bronchoscopic Navigation Using the SO‑MAC}
\label{Navigation}

\begin{figure}[t]
    \centering
    \includegraphics[ width=0.45\textwidth]{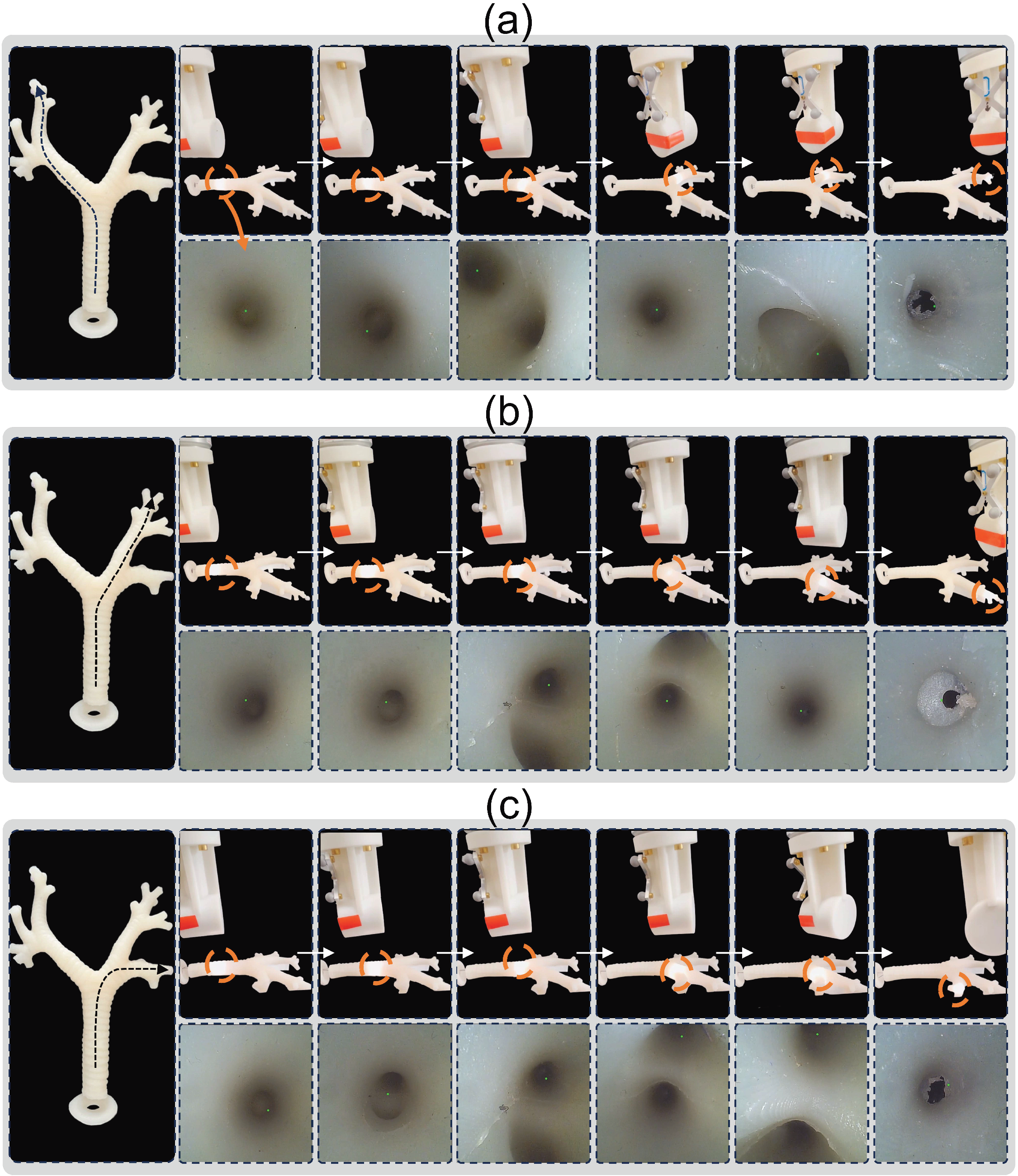}
    \caption{Bronchoscopic navigation of the SO‑MAC through three bronchial branches.
(a)–(c) Trajectories are marked by black dashed lines; the upper and lower rows show the external and endoscopic views, respectively.
Green dots indicate the detected lumen centers, and orange dashed circles mark the current SO‑MAC position.
    }
    \label{fig.exp3_procedure}
\end{figure}
A visual‑feedback, model‑free controller \cite{tan2023model} aligned the SO‑MAC tip with the target lumen center by minimizing the distance between the endoscopic image center and the detected lumen center \cite{Wang2015lumen}. 
A UR5e robot positioned the external permanent magnet on a plane $\sim$61mm above the bronchial phantom (Ningbo Lancet Medical Technology Co., Ltd., China), producing a 35–45mT field over a $\sim$10~mm workspace height. 
Over this workspace, the field was approximated as locally uniform, consistent with the uniform‑field assumption in \cite{pittiglio2023magnetic}. 
Axial propulsion was computer‑controlled, at each bifurcation, the operator selected the target lumen for alignment before continuing advancement.

We navigated the SO-MAC through three paths in the bronchial phantom (Fig. \ref{fig.exp3_procedure}(a)–(c)); each path is illustrated by six paired procedural images showing the SO-MAC motion inside the phantom and the corresponding external magnet poses. 
In our design, we assumed a clamped pivot as a conservative worst case. 
In contrast, in the bronchial lumen, contact with the compliant wall provides less restrictive support and relaxes the required turning radius; thus, pivot formation and bifurcation traversal should be no more difficult than in the clamped case.

SO-MAC successfully navigated three bronchial paths of different lengths: path (a) with a length of approximately $23.1~\rm{cm}$, path (b) of $19.5~\rm{cm}$, and path (c) of $16.2~\rm{cm}$. 
During navigation, the external magnet, positioned above the camera, followed and steered the SO-MAC through the phantom. 
Path (a) was the longest, requiring two unsupported $60^\circ$ turns at bifurcations. 
Path (c) presented the greatest challenge, as its final lumen orientation was nearly perpendicular to the trachea and lacked structural support at the bifurcations, demonstrating that the proposed steering‑pivot design provided sufficient steerability and pushability.

\section{Discussion}
\subsection{Significance of Stiffness Optimized Design}
Interventional navigation with magnetically actuated catheters requires both large-angle steerability and reliable propulsion transmission (pushability). Increasing axial/bending stiffness can improve pushability by suppressing friction- and compression-induced kinking/buckling, but excessive stiffness—especially at large steering angles—can reduce steerability, increase geometric constraint, and ultimately degrade propulsion transmission. This creates a fundamental design trade-off between structural rigidity and steerability for complex, bifurcating luminal networks.

To reconcile this trade-off, we proposed the SO‑MAC, which combines a distal-to-proximal stiffness increase with proximally concentrated bending enabled by alternating magnetic and spring segments. The spring backbone provides optimized stiffness and elastic recovery, promoting self-straightening of the alignment part to support propulsion transmission while steering is achieved by bending concentrated near a stable proximal pivot.  As a result, SO‑MAC maintains high propulsion efficiency across a wide steering range (0$^\circ$–180$^\circ$) and facilitates reliable entry from a constrained parent lumen into sharply oriented child lumens. Importantly, our recursive stiffness design clarifies how stiffness should increase to maintain the pivot location for concentrated bending and also reveals practical limits beyond which further stiffening becomes counterproductive.

\subsection{Comparison of Propulsion Transmission}
\label{Propulsion Transmission}
\begin{figure}[t]
    \centering
    \includegraphics[width=0.35\textwidth]{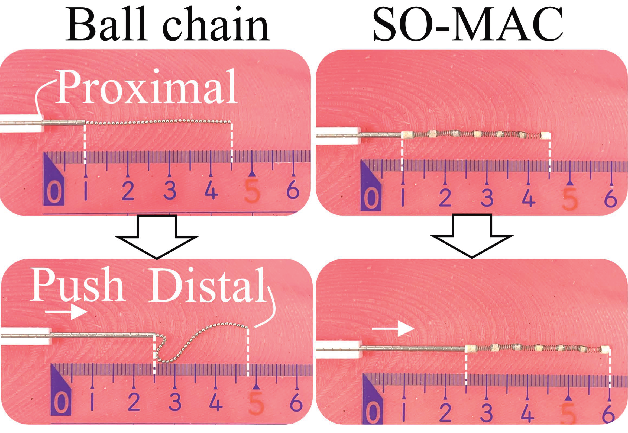}
    \caption{Comparison of propulsion transmission between the gradient-stiffness SO-MAC and a uniformly low-stiffness magnetic ball chain on a silicone substrate.}
    \label{fig:pushability}
\end{figure}
Although it is well established that a gradual increase in stiffness is essential for interventional catheters to prevent collapse\cite{lawrence2005materials}, this design principle—already widely adopted in both non‑magnetic interventional catheters \cite{lawrence2005materials,yang2025degradable} and MAC \cite{dreyfus2024dexterous}—was further verified here through a direct propulsion‑transmission comparison. 

We tested a 1\,mm‑diameter magnetic ball chain of the same total length as the SO‑MAC on a silicone substrate that mimics the frictional conditions of biological tissue. 
When the proximal end was pushed 15 mm by a rigid magnetic rod, which was linearly driven at a constant speed in a guide slot by a linear motor, the distal end of the SO‑MAC transmitted almost the entire displacement ($\approx$ 15 mm), whereas the magnetic ball chain transmitted only about 3 mm (Fig. \ref{fig:pushability}).
This result clearly demonstrates the superior propulsion‑transmission performance of the gradient‑stiffness SO‑MAC. 

The experiment reaffirms that the gradient‑stiffness design effectively prevents kinking or collapse during propulsion \cite{lawrence2005materials}. The overly soft ball chain tends to buckle under frictional loading, particularly within large lumens such as the trachea. Because bronchial lumens decrease in diameter from proximal to distal airways, the ball chain may easily collapse in wider sections despite its favorable behavior in narrow lumens \cite{pittiglio2023magnetic}.
In contrast, the SO‑MAC, endowed with inherent elasticity and gradient stiffness, self‑recovers to a straight configuration even without magnetic actuation, maintaining efficient propulsion and structural integrity across lumens of varying diameters.
The test was performed without magnetic actuation to reveal the intrinsic mechanical performance of the two catheters. Although a strong field could partially straighten the ball chain, such improvement relies on continuous magnetic assistance, whereas the SO‑MAC inherently preserves straightness and pushability, highlighting the intrinsic advantage of its gradient‑stiffness design and reduced control dependence.

\section{Conclusion}
\label{sec:conclusions}
This work presents a stiffness‑optimized, multi‑segment magnetically actuated catheter (SO‑MAC) that addresses the steerability–pushability trade-off via a gradient‑stiffness design and a steering‑pivot mechanism that concentrates bending proximally while keeping the alignment part nearly straight for axial force transmission. Experiments validated the proposed kinetostatic model and confirmed stable pivot control and effective navigation in bronchial phantoms. With a lightweight structure, a high inner‑to‑outer diameter ratio, and axially compliant spring segments, SO‑MAC is well-suited for integration with working channels and medical instruments, potentially reducing perforation risk and enabling future therapeutic functions such as targeted drug delivery.

\bibliographystyle{IEEEtran}
\bibliography{main}
\begin{IEEEbiography}[{\includegraphics[width=1in,height=1.25in,clip,keepaspectratio]{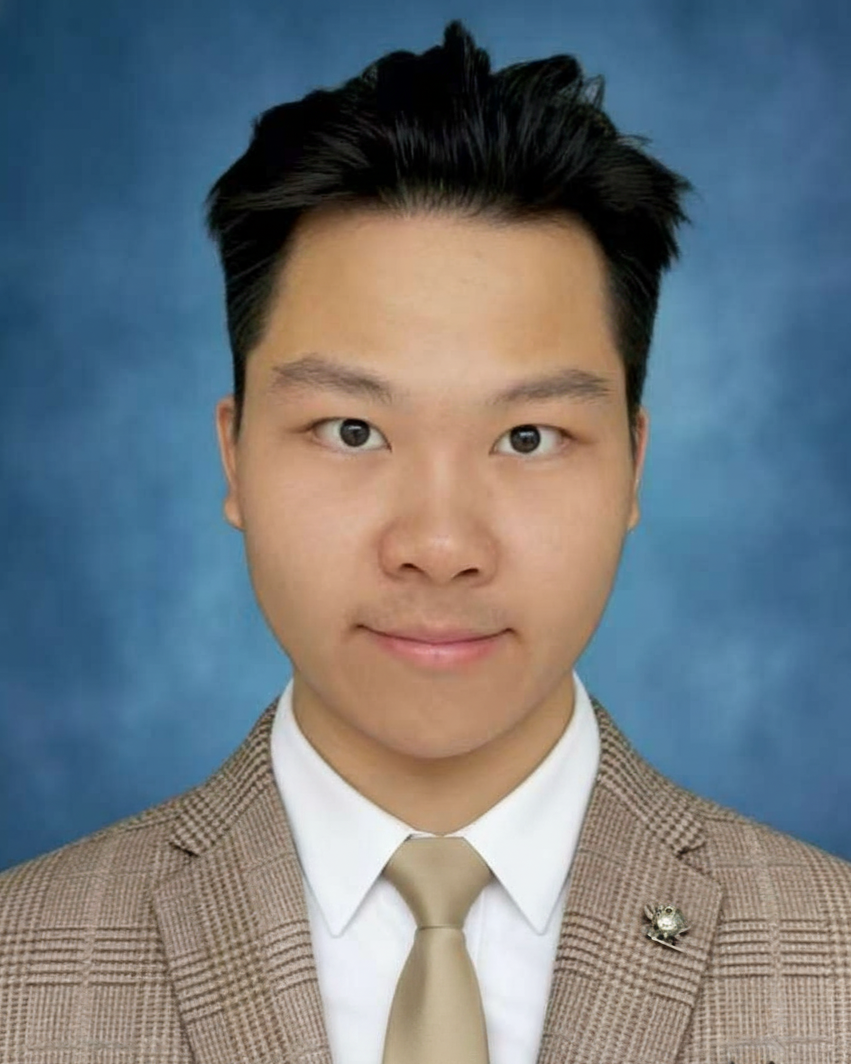}}]{Jiewen Tan}
received the Bachelor of Engineering degree in Mechanical Design, Manufacturing and Automation and the M.Eng. degree in Mechanical Engineering from the Harbin Institute of Technology (Shenzhen), Shenzhen, China, in 2021 and 2023, respectively, and is currently pursuing the Ph.D. degree in Mechanical and Automation Engineering at The Chinese University of Hong Kong, Hong Kong. His research interests include surgical robotics, magnetically actuated endoscopes, and robot modeling and control.
\end{IEEEbiography}

\begin{IEEEbiography}[{\includegraphics[width=1in,height=1.25in,clip,keepaspectratio]{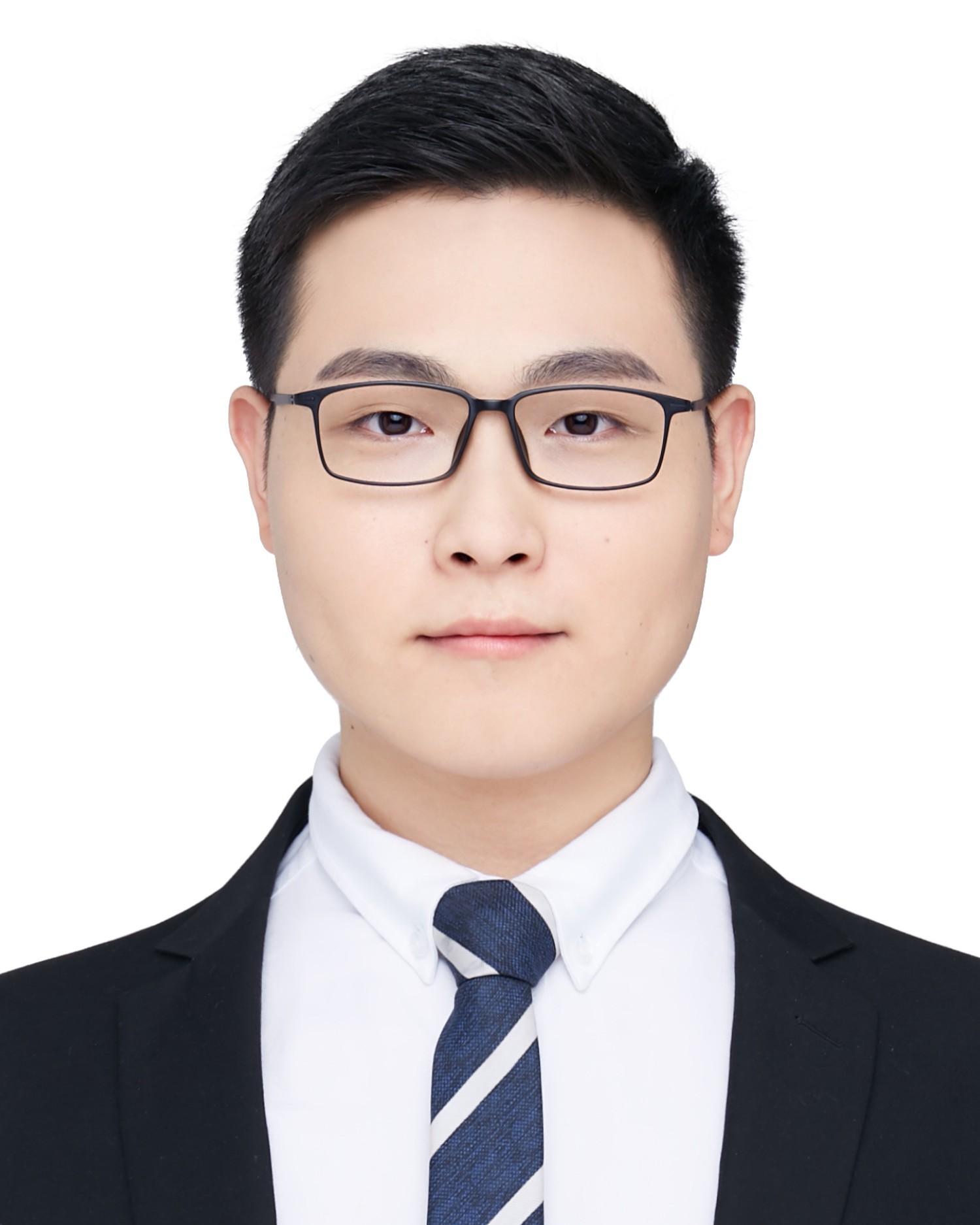}}]{Junnan Xue}
(Graduate Student Member, IEEE) received the B.E. degree in mechanical and electronic engineering from the Taiyuan University of Technology, Taiyuan, China, in 2020, and the M.E. degree in mechanical engineering from the Harbin Institute of Technology, Shenzhen, China, in 2022. He is currently working toward the Ph.D. degree in mechanical and automation engineering with The Chinese University of Hong Kong, Hong Kong. His research interests include medical robotics and small-scale robotics.
\end{IEEEbiography}

\begin{IEEEbiography}[{\includegraphics[width=1in,height=1.25in,clip,keepaspectratio]{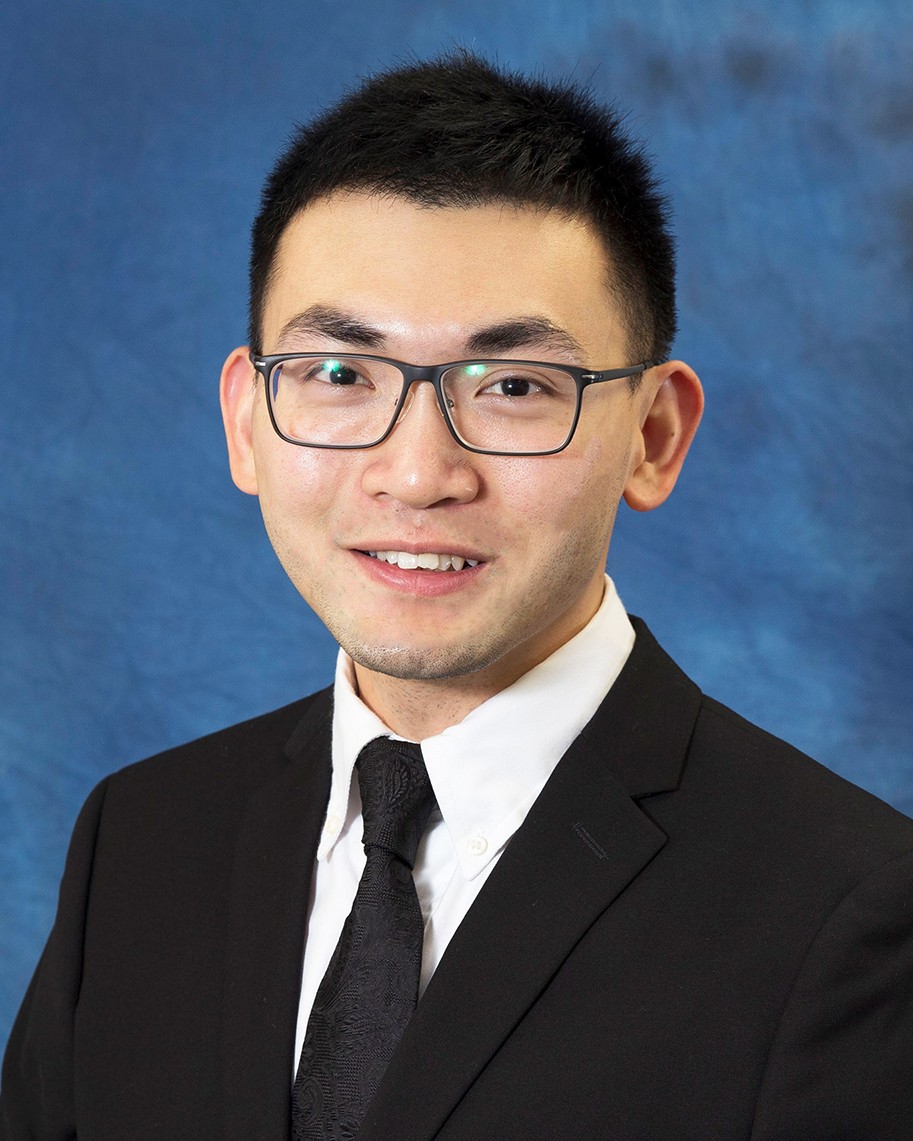}}]{Shing Shin Cheng}
(Member, IEEE) received the B.S. degree in mechanical engineering from the Johns Hopkins University, USA, in 2013, and Ph.D. degree in robotics from the Georgia Institute of Technology, USA, in 2018. He is currently an Associate Professor in the Department of Mechanical and Automation Engineering, The Chinese University of Hong Kong, Hong Kong, China. His research interests include flexible surgical robotics, image-guided surgical systems, and robot modeling and control.
\end{IEEEbiography}

\begin{IEEEbiography}	[{\includegraphics[width=1in,height=1.25in,clip,keepaspectratio]{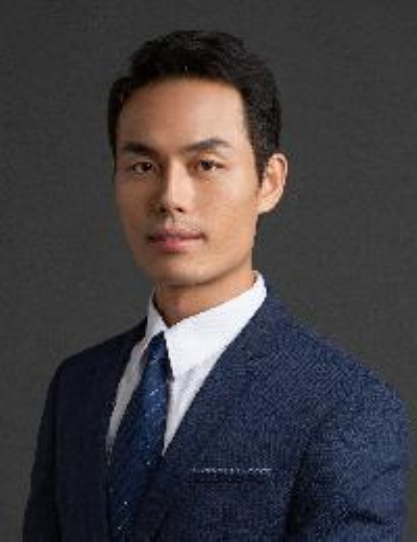}}]{Shuang Song} 
(Member, IEEE) received B.S. degree in Computer Sc. \& Tech. from North Power Electric University, M.S. degrees in computer architecture from Chinese Academy of Sciences, and his Ph.D. degree in computer application technology from University of Chinese Academy of Sciences, China, in 2007, 2010 and 2013, respectively. Now he is a Professor in Harbin Institute of Technology, Shenzhen, China. His main research interests include magnetic tracking and actuation for Bioengineering applications.
\end{IEEEbiography}

\begin{IEEEbiography}
[{\includegraphics[width=1in,height=1.25in, clip,keepaspectratio]{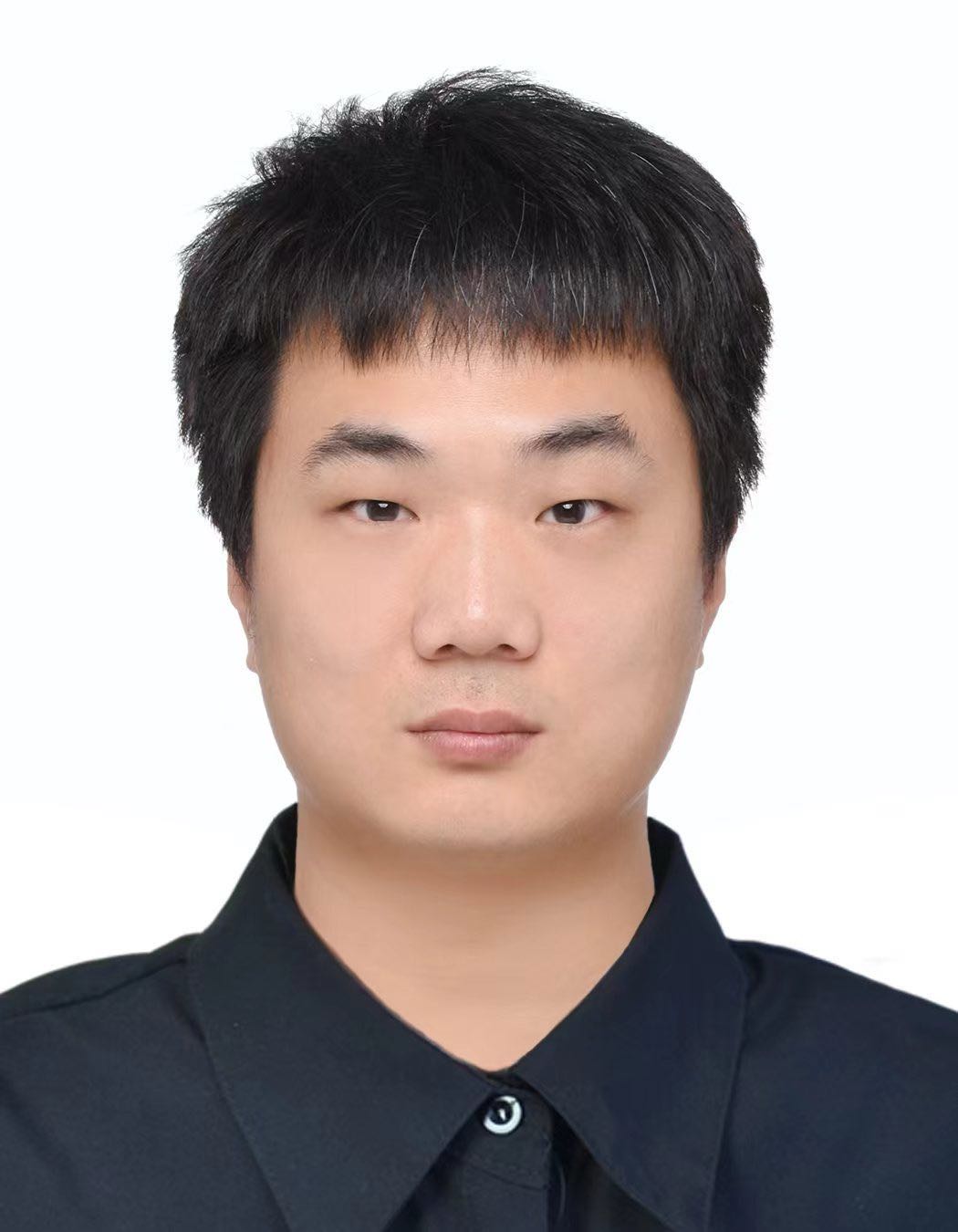}}]
{Erli Lyu} received the B.E. degree in measurement and instrumentation program from the Northeastern University at Qinhuangdao (NEUQ), Hebei, China, in 2014, and the master’s degree in control science and engineering and the Ph.D. degree in mechanical engineering from the Harbin Institute of Technology (Shenzhen), Shenzhen, China, in 2016 and 2023 respectively. He is currently a Lecturer with the Faculty of Applied Science, Macao Polytechnic University. His research interests include motion planning, manipulation, navigation, and pose estimation.
\end{IEEEbiography}
\begin{IEEEbiography}
[{\includegraphics[width=1in,height=1.25in, clip,keepaspectratio]{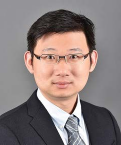}}]
{Jiaole Wang} (Member, IEEE) received the B.E. degree in mechanical engineering from Beijing Information Science and Technology University, Beijing, China, in 2007, the M.E. degree in human and artificial intelligent systems from University of Fukui, Fukui, Japan, in 2010, and the Ph.D. degree in electronic engineering from The Chinese University of Hong Kong (CUHK), Hong Kong, in 2016. He was a Research Fellow with the Pediatric Cardiac Bioengineering Laboratory, Department of Cardiovascular Surgery, Boston Children’s Hospital and Harvard Medical School, Boston, MA, USA. He is currently a Professor with the School of Biomedical Engineering, Harbin Institute of Technology, Shenzhen, China. His main research interests include medical and surgical robotics, image-guided surgery, human-robot interaction, and magnetic tracking and actuation for biomedical applications.
\end{IEEEbiography}
\end{document}